%% file: main_arxiv.tex
\documentclass[10pt,twocolumn,letterpaper]{article}

\usepackage{iccv}              %

\input{preamble}

\definecolor{iccvblue}{rgb}{0.21,0.49,0.74}
\usepackage[pagebackref,breaklinks,colorlinks,allcolors=iccvblue]{hyperref}

\def\paperID{13184} %
\def\confName{ICCV}
\def\confYear{2025}

\title{\ours: Dynamic Merging and Virtual Unmerging for Efficient VLMs}

\author{%
{Zhenhailong Wang}\textsuperscript{\textnormal{1}}\textsuperscript{*},
{Senthil Purushwalkam}\textsuperscript{\textnormal{2}}\textsuperscript{*}, 
{Caiming Xiong}\textsuperscript{\textnormal{2}}, 
{Silvio Savarese}\textsuperscript{\textnormal{2}},\\ 
{Heng Ji}\textsuperscript{\textnormal{1}},
{Ran Xu}\textsuperscript{\textnormal{2}}
\\ 
\textsuperscript{1}University of Illinois Urbana-Champaign, \textsuperscript{2}Salesforce Research\\
\small\textsuperscript{*}Equal Contribution
}

\newcommand{\ours}{\textsc{DyMU}\xspace}

\newcommand{\dtome}{DToMe\xspace} %
\newcommand{\dtomefull}{Dynamic Token Merging\xspace} %
\newcommand{\vtu}{VTU\xspace} %
\newcommand{\vtufull}{Virtual Token Unmerging\xspace} %

\definecolor{MyDarkBlue}{rgb}{0,0.08,0.8}
\definecolor{MyDarkGreen}{RGB}{45,155,45}
\definecolor{MyDarkRed}{rgb}{0.8,0.02,0.02}
\definecolor{MyOrange}{rgb}{1.0, 0.4, 0.2}
\definecolor{MyPurple}{RGB}{111,0,255}
\definecolor{MyRed}{rgb}{0.8,0.0,0.0}
\definecolor{MyGold}{rgb}{0.75,0.6,0.12}
\definecolor{MyDarkgray}{rgb}{0.66, 0.66, 0.66}

\newcommand{\uniq}[0]{\text{un}}

\everydisplay{%
  \setlength\abovedisplayskip{4pt}%
  \setlength\belowdisplayskip{4pt}%
  \setlength\abovedisplayshortskip{4pt}%
  \setlength\belowdisplayshortskip{4pt}%
}

\makeatletter
\renewcommand\paragraph{%
  \@startsection{paragraph}{4}{\z@}%
    {0.2\baselineskip}%
    {-1em}%
    {\normalfont\normalsize\bfseries}%
}
\makeatother

\begin{document}

\twocolumn[{%
\renewcommand\twocolumn[1][]{#1}%
\maketitle
\begin{center}
    \centering
    \captionsetup{type=figure}
    \vspace{-30pt}
    \includegraphics[width=0.94\textwidth]{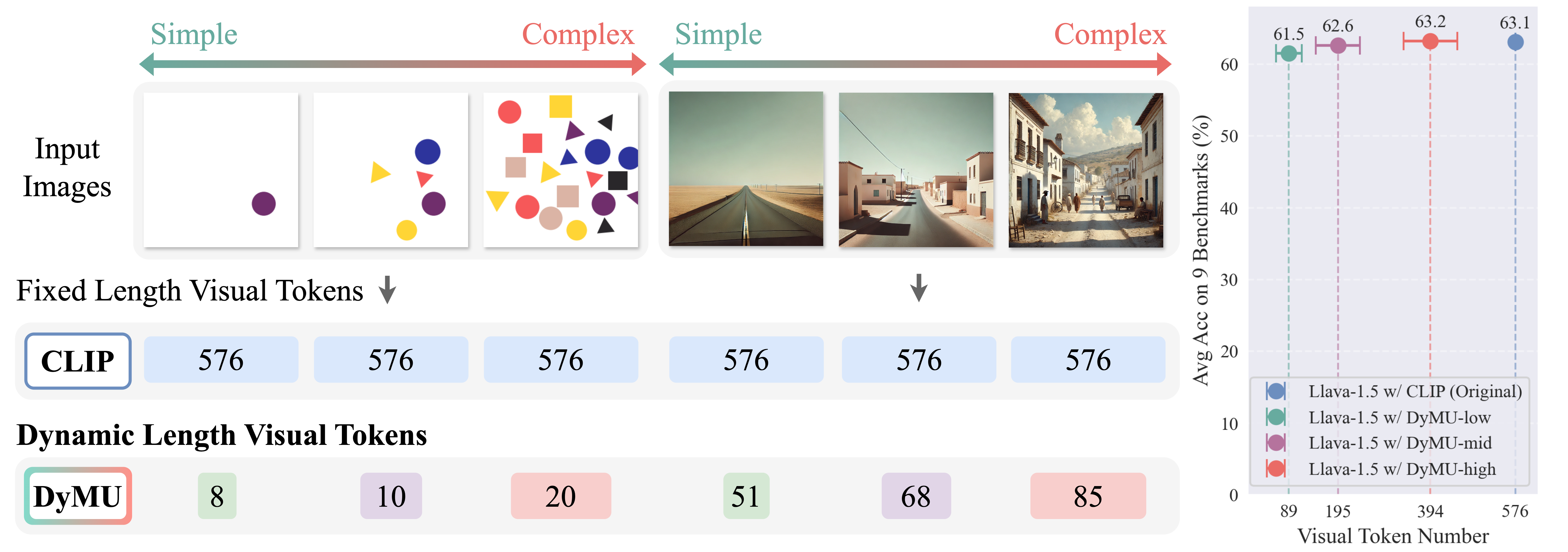}
    \vspace{-3pt}
    \caption{\textbf{Dynamic Merging and Virtual Unmerging} (DyMU) adaptively reduces visual token lengths based on \textit{image complexity}, as shown on the left where simpler images are represented using fewer tokens. In contrast, existing representations (like CLIP~\cite{clip}) always use the same number of tokens regardless of image content. Note that this limitation also exists in dynamic resolution encoders~\cite{llava-ov,Qwen2VL}, where token length depends solely on image dimensions rather than content.
    DyMU applied to recent VLMs (right) maintains competitive performance across different token compression levels. This training-free approach preserves key semantic information, offering a more efficient plug-and-play alternative to VLMs with fixed-length visual tokens.
    }
    \label{fig:teaser}
\end{center}%
}]

\input{arxiv_sections/0_abstract}

\input{arxiv_sections/1_intro}

\input{arxiv_sections/4_related_work}

\input{arxiv_sections/2_method}

\input{arxiv_sections/3_experiments}
\input{arxiv_sections/5_conclusion}

{
    \small
    \bibliographystyle{ieeenat_fullname}
    \bibliography{main}
}

\clearpage

\input{arxiv_sections/appendix}

\end{document}

%% file: preamble.tex
\usepackage{xcolor}
\usepackage{color}
\usepackage{epsfig}
\usepackage{graphicx}

\usepackage{adjustbox}
\usepackage{array}
\usepackage{booktabs}
\usepackage{colortbl}
\usepackage{wrapfig}
\usepackage{hhline}
\usepackage{multirow}
\usepackage{wrapfig}
\usepackage{floatflt}

\usepackage{tabularx} %
\usepackage{makecell}
\usepackage{wrapfig}

\usepackage{bm}
\usepackage{nicefrac}

\usepackage{changepage}
\usepackage{extramarks}
\usepackage{fancyhdr}
\usepackage{lastpage}
\usepackage{setspace}
\usepackage{soul}
\usepackage{xspace}

\usepackage{amsmath}

\usepackage{algorithm}
\usepackage{algpseudocode}
\usepackage{enumerate}
\usepackage{enumitem}  %
\usepackage{ulem} %

\usepackage{makecell}

\usepackage{pifont} %
\usepackage{caption}

\usepackage[symbol]{footmisc}

\usepackage[most]{tcolorbox}

\newcommand{\addcheckemoji}{\raisebox{-.2\height}{\includegraphics[height=12pt]{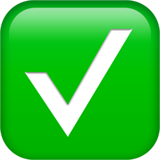}}}
\newcommand{\addcrossemoji}{\raisebox{-.2\height}{\includegraphics[height=12pt]{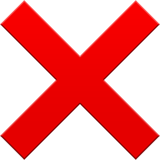}}}

%% file: arxiv_sections/0_abstract.tex
\begin{abstract}

\vspace{-10pt}

We present \ours, an efficient, training-free framework that dynamically reduces the computational burden of vision-language models (VLMs) while maintaining high task performance. Our approach comprises two key components. First, Dynamic Token Merging (\dtome) reduces the number of visual token embeddings by merging similar tokens based on image complexity, addressing the inherent inefficiency of fixed-length outputs in vision transformers. Second, Virtual Token Unmerging (VTU) simulates the expected token sequence for large language models (LLMs) by efficiently reconstructing the attention dynamics of a full sequence, thus preserving the downstream performance without additional fine-tuning. Unlike previous approaches, our method dynamically adapts token compression to the content of the image and operates completely training-free, making it readily applicable to most state-of-the-art VLM architectures. Extensive experiments on image and video understanding tasks, demonstrate that \ours can reduce the average visual token count by 32\%-85\% while achieving comparable performance to full-length models, across diverse VLM architectures, including the recently popularized AnyRes-based visual encoders. Furthermore, through qualitative analyses we demonstrate that \dtome effectively adapts token reduction based on image complexity, and unlike existing systems, provides users more control over computational costs. 
Project page: \url{https://mikewangwzhl.github.io/dymu}.

\end{abstract}

%% file: arxiv_sections/1_intro.tex
\vspace{-12pt}
\section{Introduction}
\label{sec:intro}

Recent large vision-language models (VLMs) have demonstrated breakthroughs in computer vision tasks such as image captioning~\cite{mscoco_caption}, open-vocabulary object detection~\cite{gupta2019lvis}, visual-question answering~\cite{vqav2} and OCR~\cite{ocrvqa} by leveraging the reasoning capabilities of large language models (LLMs) to enhance visual understanding. 
Most VLMs follow a common approach: a visual encoder extracts features from images or videos and projects them into the same embedding space as textual features. These visual embeddings are then processed by LLMs alongside textual query features, enabling complex understanding and reasoning tasks while directly benefiting from advancements in LLM capabilities. 

As expected, the quality of the final predictions from the LLM relies heavily on the richness of the visual features and the amount of semantic detail captured by the encoder. Consequently, research has focused on improving visual encoders to extract increasingly fine-grained features, leading to architectures that can capture intricate details.
However, this level of detail comes at a cost --- the computational burden during training and inference.

To process high-resolution images while preserving fine-grained details, modern visual encoders generate a large number of tokenized representations. Furthermore, state-of-the-art VLMs like LLaVA-OneVision~\cite{llava-ov} and Qwen-2.5VL~\cite{qwen25vl} use vision transformers (ViTs) that scale the number of tokens with the resolution of the image or number of frames in videos. For example, the visual encoder in LLaVA-OneVision would produce 9477 tokens for an image of 1280$\times$960 resolution.
In contrast, the number of tokens in the textual queries for vision tasks is relatively low. On common benchmarks that represent real world use cases, textual queries often consist of just a few tokens, e.g., $\sim$24 on MME~\cite{fu2023mme}. This stark contrast highlights that the computational burden of processing vision tasks generally arises primarily from the large number of visual tokens.

We first make an interesting observation: in current visual encoders, the number of tokens generated for an image does not depend on the content of the image. In Figure~\ref{fig:teaser}, we illustrate this with some examples --- a CLIP~\cite{clip} representation leads to the same embedding size on a blank image with a small circle and on a complex scene depicting buildings, vehicles and people. In contrast, textual tokens are more closely tied to the amount of content conveyed --- more words are required to describe more information.
An average sentence length in English is around 15–20 words~\cite{francis79browncorpus}, meaning that \textit{regardless of the content of the image}, the language model in LLaVA-OneVision~\cite{llava-ov} has to process the equivalent of 400-500 sentences for each high-resolution image.

In this work, we propose \textbf{Dynamic Merging and Virtual Unmerging (\ours)}, which comprises two key methods for modifying existing pre-trained Vision-Language Models (VLMs). First, we introduce Dynamic Token Merging (Sec~\ref{subsec:batchtome}), which allows the \textit{visual encoder} to generate variable-length token sequences based on the complexity of the image. Second, we present Virtual Token Unmerging (Sec~\ref{subsec:tokexpansion}), enabling the \textit{LLM decoder} to process shorter dynamic visual token sequences while efficiently approximating the full-length sequence. Crucially, we demonstrate that both of these modifications \textit{do not require additional fine-tuning of the pre-trained VLM}. Furthermore, Dynamic Token Merging is compatible with any Vision Transformer (ViT)-based visual encoder, and Virtual Token Unmerging can be applied to any LLM that utilizes Rotary Position Embedding (RoPE)~\cite{rope}.

We show that VLMs modified with our methods can maintain the performance of the full model while reducing the average token count by 32\%-85\% (see Sec~\ref{sec:quant}). In addition to improving efficiency, our approach offers users greater control over token costs compared to existing systems (e.g., GPT-4o), which incur a fixed token count per image based solely on resolution. In Sec~\ref{sec:qualitative}, we demonstrate example applications on how the number of visual tokens can be further reduced by combining \ours with pre-processing tools such as background removal, object detection, etc. Through comprehensive quantitative experiments (Sec~\ref{sec:quant}), we verify that our method works effectively across different VLM architectures, with varying pre-training strategies, visual encoders, and training datasets.

%% file: arxiv_sections/4_related_work.tex
\section{Related Work}

\input{tables/comparison_w_prev_work}

\paragraph{Efficient Vision-Language Models}
Recent efforts in large vision-language models (VLMs) have primarily focused on reducing computational overhead during the pre-filling and VLM decoding phases. That is, given a full sequence of visual tokens from a visual encoder, such as CLIP, these approaches perform token pruning and merging~\cite{llava-prumerge, fast-v, xing2024pyramiddrop, zhang2024sparsevlm, dynamic-llava, dynamic_token_reduction, ye2024atp}, distillation~\cite{ye2024voco}, or resampling~\cite{llama-vid, MQT-LLAVA, li2024tokenpacker, llava_mini} to improve efficiency in either the projectors or the VLM decoder blocks.
However, we identify several key limitations: (1) Most existing methods, including all training-free approaches~\cite{fast-v, xing2024pyramiddrop, zhang2024sparsevlm} predefine a fixed compression ratio for any input image regardless of its complexity. While \cite{ye2024atp} proposed an adaptive token pruning framework that enables variable-length compression, it requires retraining the backbone VLM with additional modules. Such training can be costly or infeasible as mainstream VLMs rarely open-source their full training recipe and data. (2) All existing methods retain a frozen, fixed-length visual encoder, overlooking the potential for further efficiency improvements within the visual encoder itself.
In this work, we aim to explore a simple training-free algorithm for variable length visual token compression, which can be directly applied to cutting-edge VLM architectures including Any-Resolution models and RoPE embeddings.

\paragraph{Efficient Vision Transformers}
We also draw inspiration from a separate line of research~\cite{token_pooling, tome, wu2023ppt, Avit, adaptive_sparse_vit} aimed at improving the efficiency of Vision Transformers (ViTs) themselves, which is still the main go-to architecture for visual encoders~\cite{clip, siglip, oquab2023dinov2}. In particular, ToMe~\cite{tome} merges a predefined number of tokens within each ViT block using bipartite soft matching. However, the effectiveness of such methods in coordination with VLM backbones remains largely unexplored. Our experiments in \S\ref{sec:experiments} show that naively applying ToMe to visual encoders in pretrained VLMs results in a significant drop in performance. To address this issue, we further propose an efficient ``virtual unmerging'' algorithm to boost the performance of VLMs without training with the modified encoders that output reduced token numbers.

%% file: tables/comparison_w_prev_work.tex
\begin{table}[th!]
\centering
\renewcommand{\arraystretch}{1.2} %
\resizebox{\columnwidth}{!}{
\rowcolors{2}{gray!10}{white}
\begin{tabular}{@{}l >{\centering\arraybackslash}m{1.6cm} >{\centering\arraybackslash}m{1.05cm} >{\centering\arraybackslash}m{1.2cm} >{\centering\arraybackslash}m{1.15cm} >{\centering\arraybackslash}m{1.4cm} >{\centering\arraybackslash}m{1.0cm} @{}}
    \toprule
        & \textbf{\begin{tabular}[c]{@{}c@{}}Component \\ Improved \end{tabular}} &
        \textbf{\begin{tabular}[c]{@{}c@{}}Dynamic \\ Length \end{tabular}} &
        \textbf{\begin{tabular}[c]{@{}c@{}}No Addn. \\ Modules \end{tabular}} &
        \textbf{\begin{tabular}[c]{@{}c@{}}Training \\ Free \end{tabular}} &
        \textbf{\makebox[1.4cm][c]{\begin{tabular}[c]{@{}c@{}}Granularity \\ Control\end{tabular}}} &
        \textbf{\begin{tabular}[c]{@{}c@{}}Extra \\ Cond. \end{tabular}} \\
    \midrule
        LLaMA-VID~\cite{llama-vid}  & Projector  & \addcrossemoji{} & \addcrossemoji{} & \addcrossemoji{} & \addcrossemoji{} & None \\  
        Fast-V~\cite{fast-v}     &  Decoder   & \addcrossemoji{} & \addcheckemoji{} & \addcheckemoji{} & \addcheckemoji{} & None \\  
        SparseVLM~\cite{zhang2024sparsevlm}  & Decoder   & \addcheckemoji{} & \addcheckemoji{} & \addcheckemoji{} & \addcheckemoji{} & \textcolor{red}{Text} \\  
        MQT-LLaVA~\cite{MQT-LLAVA}  & Projector  & \addcrossemoji{} & \addcrossemoji{} & \addcrossemoji{} & \addcrossemoji{} & None \\  
        LLaVA-Prumerge~\cite{llava-prumerge} & Projector  & \addcrossemoji{} & \addcheckemoji{} & \addcrossemoji{} & \addcheckemoji{} & None \\  
        TokenPacker~\cite{li2024tokenpacker}    & Projector  & \addcrossemoji{} & \addcrossemoji{} & \addcrossemoji{} & \addcheckemoji{} & None \\  
        ATP-LLaVA~\cite{ye2024atp}      & Decoder   & \addcheckemoji{} & \addcrossemoji{} & \addcrossemoji{} & \addcheckemoji{} & \textcolor{red}{Text} \\  
        LLaVA-mini~\cite{llava_mini}         & Projector  & \addcrossemoji{} & \addcrossemoji{} & \addcrossemoji{} & \addcrossemoji{} & None \\  
        \textbf{\ours{}}                     & \textbf{Encoder} \& Decoder & \addcheckemoji{} & \addcheckemoji{} & \addcheckemoji{} & \addcheckemoji{} & None \\  
    \bottomrule
\end{tabular}
}

\label{tab:comparison_w_prev_work}
\end{table}

%% file: arxiv_sections/2_method.tex
\input{figures/method_figure}

\section{Method}
\label{sec:method}

In this section, we present the main technical details of proposed method. In Section~\ref{subsec:batchtome}, we present our proposed Dynamic Token Merging (\dtome) --- a training-free method to dynamically reduce the number of output tokens by visual encoders based on the complexity of the image content. 
In Section~\ref{subsec:tokexpansion}, we introduce Virtual Token Unmerging (\vtu) --- an approach to process the reduced visual tokens through the language model while efficiently simulating the standard number of visual tokens. This method utilizes the tracked positions of the redundant tokens to recreate a full attention matrix of the original length. 

The combination of both methods is referred to as \ours, short for Dynamic Merging and Virtual Unmerging. We illustrate the core idea in Figure~\ref{fig:method}.
\ours can be applied to any VLM that uses transformer-based visual encoders and RoPE-based transformer language models. The proposed modifications to the architecture do not introduce any additional learnable parameters and most importantly, do not require any additional fine-tuning of the VLM.

\subsection{Dynamic Token Merging (\dtome)}
\label{subsec:batchtome}

Most recent large vision-language models (VLMs) use vision transformers (ViTs) like CLIP~\cite{clip} or SigLIP~\cite{siglip} to encode images into a sequence of visual tokens. For a fixed resolution input image, the ViT architecture always outputs the same number of token embeddings, leading to low efficiency in VLMs.
Our approach draws inspiration from ToMe\cite{tome}, a prior work which reduces the number of output tokens to a predefined fixed number. 
However, predefining the reduction ratio can still lead to a misalignment between the information of an image and the number of tokens needed for representing it. 

Here we propose Dynamic Token Merging (\dtome), an extension of ToMe that adaptively merges similar tokens in ViT layers, ensuring the output token count aligns with image complexity. \dtome merges tokens based on a similarity threshold while maintaining a record of merged tokens to ensure their influence is properly propagated through subsequent transformer layers. 
To find the thresholds, we propose a inference-only batch-level bipartite merging algorithm which leverages the natural variance of image complexity in randomly sampled images.

\paragraph{Identifying Redundant Tokens} Let us represent the output of the self-attention layer in the ViT layer $i$ as $x_{i}\in \mathbb{R}^{N_i\times D}$, where $N_i$ is the sequence length\footnote{For standard ViT without any merging, $N_i$ is constant across layers} and $D$ is the embedding dimension. Similarly, let the keys computed in the self-attention layer be represented by $k_i\in \mathbb{R}^{N_i\times D_k}$. In each transformer block, we apply an additional \dtome operator to $x_i$. Drawing inspiration from \cite{tome}, we follow a bipartite soft matching strategy to identify which tokens need to be merged. First, we divide the $N_i$ tokens into two sets (say $\mathbb{A}$ and $\mathbb{B}$) by assigning alternating tokens in sequence them. We then compute a bipartite assignment between the two sets of tokens by assigning token $t\in \mathbb{A}$ to $t_B = \textstyle\arg\max\limits_{n \in \mathbb{B}} (k_i[t]^T~k_i[n])$ (token with the most similar key). This gives us $|\mathbb{A}|$ edges with scores $\textstyle S_i[t] = (k_i[t]^T~k_i[t_B])$ for $t\in \mathbb{A}$. We then apply a threshold $\tau_i$ to retain edges $t\xrightarrow[]{} t_B$ where $S_i[t]>\tau_i$. Unlike \cite{tome}, this thresholding operation leads to a variable number of retained edges depending on the amount of redundancy demonstrated in the key embeddings $k_i$. We describe our approach for computing the thresholds below.

\paragraph{Tracking and Merging Tokens} For each token in the sequence, $x_{i}[t]$, we also track the set of positions of the tokens that have already been merged into it.  $\mathbf{P}_i[t]\subset \{1,2,\dots,N_1\}$. For each of the edges between chosen redundant tokens $t\xrightarrow[]{} t_B$, we compute merged token embeddings and the corresponding position sets as:

{\footnotesize
\vspace{-9pt}
\begin{align}
        x_i{\left[t_B\right]} &\xleftarrow{} \frac{x_i{\left[t\right]} \cdot |\mathbf{P}_i{\left[t\right]}| + x_i{\left[t_B\right]} \cdot |\mathbf{P}_i{\left[t_B\right]}|}{|\mathbf{P}_i{\left[t\right]}| + |\mathbf{P}_i{\left[t_B\right]}|} \\[10pt]
        \mathbf{P}_i{\left[t_B\right]} &\xleftarrow{} \mathbf{P}_i{\left[t_B\right]} \cup \mathbf{P}_i{\left[t\right]} \\[10pt]
        \mathbf{P}_i{\left[t\right]} &\xleftarrow{} \varnothing
\end{align}
}
Intuitively, the representation of token $t_B$ is updated to the average of $x_i[t_B]$ and $x_i[t]$, weighted by their corresponding merged position set sizes, $\mathbb{P}_i[t_B]$ and $\mathbf{P}_i[t]$. The token $t$ is then dropped since it has been merged with $t_B$, thereby reducing the token count in the next layer.

\paragraph{Finding Redundancy Thresholds} The layer-wise thresholds $\tau_i$ play a crucial role in determining how many tokens are merged. In order to determine the thresholds, we rely of statistics from a large dataset of images. First, we choose a hyper-parameter $r_i$ for each layer $i$ which represents the number of edges we expect to merge in a layer \textbf{\textit{on average across images of all complexities}}. The final output would then be expected to have an \textit{average} of $N-\sum r_i$ tokens. Using a dataset of images, we collect large batches of size $B$ which are used to perform forward computation through the layers of the ViT sequentially. For each layer, we compute the $B$ bipartite matching token edge score maps $S^{(b)}[t]$ where $b\in \{1, 2,\dots ,B\}$ as previously described. We then find the threshold $\tau_i$ as: 

{\footnotesize
\vspace{-7pt}
\begin{align}
\tau_i = \max \left\{ \tau \mid \sum^B_{b=1} \sum_{t\in \mathbb{A}^{(b)}} \mathbb{I}\left( S^{(b)}[t] > \tau \right) = B*r_i \right\}
\end{align}
}

In words, this finds the largest threshold such that $B*r_i$ tokens are merged across the batch of images. It is important to note that the number of tokens merged in each image will not necessarily be equal to $r_i$ but the \textit{average} number of tokens merged per image will be $r_i$. 
Intuitively, since the ranking of edges is over the entire batch, simpler images that have more redundant tokens will be merged more.
This process is done sequentially for each layer while only passing the remaining tokens to the next layer to obtain thresholds for every layer. We then average the layer-wise thresholds across several batches to ensure that they reflect the statistics across a diverse set of images. See Figure~\ref{fig:method} (left) for an illustration of the proposed batch-level threshold finding.

\paragraph{Size Weighted Self-attention} To ensure that the self-attention layers weigh each token based on the number of tokens that were previously merged into it, we adopt the idea of size-weighted self-attention from ToME\cite{tome} where the attention is computed as:

{\footnotesize
\vspace{-10pt}
\begin{align}
    \mathbf{A} &= \texttt{Softmax} \left(\frac{\mathbf{QK}^T}{\sqrt{d}} + \log \left[\begin{matrix} |\mathbb{P}_i[1]| \\ \vdots\\ |\mathbb{P}_i[N_i]| \end{matrix}\right]\right)
\end{align}
}

\subsection{Virtual Token Unmerging (\vtu)}
\label{subsec:tokexpansion}

The language model (LLM) in a pre-trained VLM is trained to operate on a fixed number of embeddings for each image\footnote{AnyRes\cite{llava-ov} leads to multiple fixed length embeddings}. When Dynamic Token Merging is applied to a visual encoder, this disrupts the optimized VLM and leads to a significant drop in performance (see Sec~\ref{sec:experiments}). 
In this section, we present an approach to circumvent this issue while still benefiting from processing fewer number of visual embeddings. 
Our proposed approach, \vtufull{} (\vtu), can be easily applied to any mainstream LLM that uses a RoPE~\cite{rope}-based transformer architecture. 

Consider the general case of a sequence of $N$ embeddings $e\in \mathbb{R}^{N\times D}$ of which only $N_\uniq<<N$ rows are unique. Let $e_\uniq\in \mathbb{R}^{N_\uniq\times D}$ be the unique embeddings and $M\in \{0,1\}^{N\times N_\uniq}$ be a mapping such that $e = M ~ e_\uniq$. Here $M$ is a sparse matrix with one-hot rows\footnote{Note that due to the sparsity of $M$, the time complexity of multiplying $MD, DM, M^TD, DM^T$ are all $O(NK)$ if $D$ is a dense matrix with dimensions $N\times K$ or $K\times M$.}. We now ask the question --- for various operators $f$ in an LLM, can we approximate $f(e)$ using some efficient function of $e_\uniq$ and $M$?

\paragraph{Sequence-independent Operators} 
For any operator $f$ that processes each sequence location independently, we can express 
$f(e)$ as $f(e) = M~ f(e_\uniq)$ by definition. 
This means that we only need to apply $f$ to the unique embeddings $e_\uniq$, significantly reducing computational cost while preserving the original outputs.
Many key components of modern LLMs fall into this category, including Linear layers, Activation functions (ReLU, GeLU, etc.), and Layer Normalization (along the embedding dimension $D$). The overall complexity of the MLP layers is reduced from $O(N D^2)$ to $O(N_\uniq D^2)$, resulting in a linear speedup with $N_\uniq << N$.

\subsubsection*{Virtual Unmerging for Self-Attention with RoPE}
A common layer in recent LLMs is the Self-Attention operation with Rotary Position Embedding (RoPE). Unlike sequence-independent operators, self-attention considers pairwise interactions between embeddings and assigns a unique position to each of the $N$ locations in $e$. Consequently, directly applying $f(e_\uniq)$ fails to capture the structure of $e$, generally leading to significant discrepancies in the output.  

To address this, we provide a theoretical derivation of an efficient method to compute $f(e)$ while preserving the benefits of token reduction. The key insight is to reconstruct the self-attention matrix without explicitly expanding the token sequence. We leverage the linearity of the RoPE transformation to efficiently simulate the appropriate repetitions and the positions of the unique embeddings, significantly reducing computational overhead while maintaining consistency with the full sequence computation.  

Let $Q=W_q e$, $K=W_k e$ and $V=W_v e$ be the full query, key and value matrices. Similarly, $Q_\uniq,K_\uniq$ and $V_\uniq$ are the unique queries, keys and values satisfying the mapping $M$ defined above. The RoPE Self-Attention similarity matrix is computed as $A = \mathtt{RoPE}(Q)\mathtt{RoPE}(K)^T$.

For simplicity, let us consider the case where $D=2$, so that we can write $Q = [Q_1, Q_2]$ where $Q_1, Q_2\in \mathbb{R}^{N}$. We will follow a similar notation for all queries, keys and values.  This allows us to express each query and key as a complex number \textit{i.e.} $Q[n] = Q_1[n] + \mathbf{i} Q_2[n]$. Let $\boldsymbol{\theta}\in [0, 2\pi)^N$ be the rotation angle associated with each position for RoPE. For positions $n,m\in {1,2,\dots N}$, the RoPE-based similarity~\cite{rope} is defined as:

{\footnotesize
\vspace{-5pt}
\begin{align}
A[m,n] &= \text{Re}\big( ~~e^{i\boldsymbol{\theta[m]}}Q[m]  ~~~\overline{e^{i\boldsymbol{\theta[n]}}K[n]} ~~ \big)\\
\small  &= \text{Re}\big( ~~Q[m]\overline{K[n]}  ~~ e^{i(\boldsymbol{\theta}[m]-\boldsymbol{\theta}[n])} ~~\big)
\end{align}
}

\noindent where $\overline{x},\text{Re}(x)$ denote the complex conjugate and the real part of $x$ respectively. This can be expanded as:

{\scriptsize
\vspace{-5pt}
\begin{align}
\begin{aligned}
\label{eq:attention_entry}
A[m,n] &= (Q_1[m]K_1[n] + Q_2[m]K_2[n]) \cos(\boldsymbol{\theta}[m]-\boldsymbol{\theta}[n]) \\
&\quad + (Q_1[m]K_2[n] - Q_2[m]K_1[n]) \sin (\boldsymbol{\theta}[m]-\boldsymbol{\theta}[n])
\end{aligned}
\end{align}
}

We also have the trigonometric identities:

{\footnotesize
\vspace{-5pt}
\begin{align}
\begin{aligned}
\label{eq:trig_expansion}
\cos(\theta[m] – \theta[n]) = \cos (\theta[m]) \cos(\theta[n]) + \sin(\theta[m]) \sin(\theta[n])\\
\sin(\theta[m] – \theta[n]) = \sin (\theta[m]) \cos(\theta[n]) - \cos(\theta[m]) \sin(\theta[n])
\end{aligned}
\end{align}
}

Let $C=\mathtt{diag}(\cos(\boldsymbol{\theta})), S=\mathtt{diag}(\sin(\boldsymbol{\theta}))$. Using Eq ~\ref{eq:attention_entry} \& ~\ref{eq:trig_expansion}, the matrix form for self-attention similarities is:

{\footnotesize
\vspace{-10pt}
\begin{align}
\begin{aligned}
A &= C QK^{\top} C + S QK^{\top} S \notag + S (Q\times K^{\top})C -  C(Q\times K^{\top})S 
\end{aligned}
\end{align}
}

\noindent where {\small$QK^T = Q_1K_1^T + Q_2K_2^T,\; Q\times K^{\top} = Q_1K_2^{\top} - Q_2K_1^{\top}$}.
This formulation can be applied to queries and keys of any dimension $D$ by repeating this for the $(D/2)$ complex numbers obtained by dividing the representation into two parts. In practice, a different $\boldsymbol{\theta}$ is used for each of the $(D/2)$ components.

\input{tables/flops}

\input{tables/main_result_sota}

\input{tables/siglip_results}

Using this formulation and the mapping M, we can rewrite the attention matrix in terms of the unique queries and keys as:

{\footnotesize
\vspace{-10pt}
\begin{align}
\label{eq:expand_sim}
A &= C M Q_\text{un} K_\text{un}^{\top} M^{\top} C 
   + S M Q_\text{un} K_\text{un}^{\top} M^{\top} S \notag \\
  &\quad + S M (Q_\text{un} \times K_\text{un}^{\top}) M^{\top} C 
   - C M (Q_\text{un} \times K_\text{un}^{\top}) M^{\top} S
\end{align}
}

Observe {\small $CM, M^TC, SM, M^TS$} are highly sparse, each with at most $N$ non-zero entries. These matrices can also be pre-computed and reused across all self-attention layers. Computing $Q_\text{un} K_\text{un}^{\top}$ and $Q_\text{un}\times K_\text{un}^{\top}$ incurs an $O(N_\uniq^2)$ cost whereas the each of the other matrix multiplications in Eq~\ref{eq:expand_sim} can be efficiently computed using sparse matrix operations in $O(N N_\uniq)$.
We can then use the attention matrix to compute the final output of the layer as:

\begin{footnotesize}
\vspace{-5pt}
\begin{align}
f(e) = \mathtt{smax} ( \frac{A}{\sqrt{D}} ) V = [\mathtt{smax} ( \frac{A}{\sqrt{D}} ) M] V_\uniq \notag
\end{align}
\end{footnotesize}

Unfortunately, the output $f(e)\in \mathbb{R}^{N\times D}$ will not necessarily exhibit the same redundancy as $e$. This in turn means that the future self-attention layers cannot benefit from the efficiency of virtual token unmerging. In order to remedy this, before passing the output to the future layers, we re-introduce the redundancy by averaging the embeddings in the positions that were originally equal. We denote this \textit{re-merged} output by $f'(e_\uniq, M)$ which can be written as:

{\footnotesize
\vspace{-5pt}
\begin{align}
\label{eq:uniq_selfattn_out}
f'(e_\uniq, M) &= (M^{\top} M)^{-1} M^T f(e) \notag \\
&= (M^{\top} M)^{-1} M^T \mathtt{smax} ( \frac{A}{\sqrt{D}} ) V 
\end{align}
}

While the above averaging operation breaks the exactness of the future operations, we observe empirically (see Section~\ref{sec:experiments}) that this re-merging of tokens, that are known to be redundant, causes minimal drop in performance.  

\paragraph{Overall Efficiency} The computation of attention matrix $A$ incurs a cost of $O(N_{un}^2 D + N N_{un} D)$ (due to the $D/2$ components). Followed by the softmax and sparse matrix multiplications in Eq~\ref{eq:uniq_selfattn_out} which incur a cost of $O(N^2 + N_\uniq^2 D)$. Therefore, the overall complexity for RoPE Self-Attention with Virtual Token Unmerging is $O(N_\uniq N D)$. For comparison, the full RoPE Self-Attention on a sequence length of $N$ would be an $O(N^2 D)$ operation. 
Therefore, in theory, efficiency improves at least linearly with the number of redundant tokens in terms of FLOPs. Table \ref{tab:flops} shows the FLOPs comparison for the attention block. 
In practice, we find that the wall-clock time difference is marginal due to PyTorch’s highly optimized attention and dense matrix-multiplication implementations.

%% file: figures/method_figure.tex
\begin{figure*}[ht]
    \centering
    \includegraphics[width=0.9\linewidth]{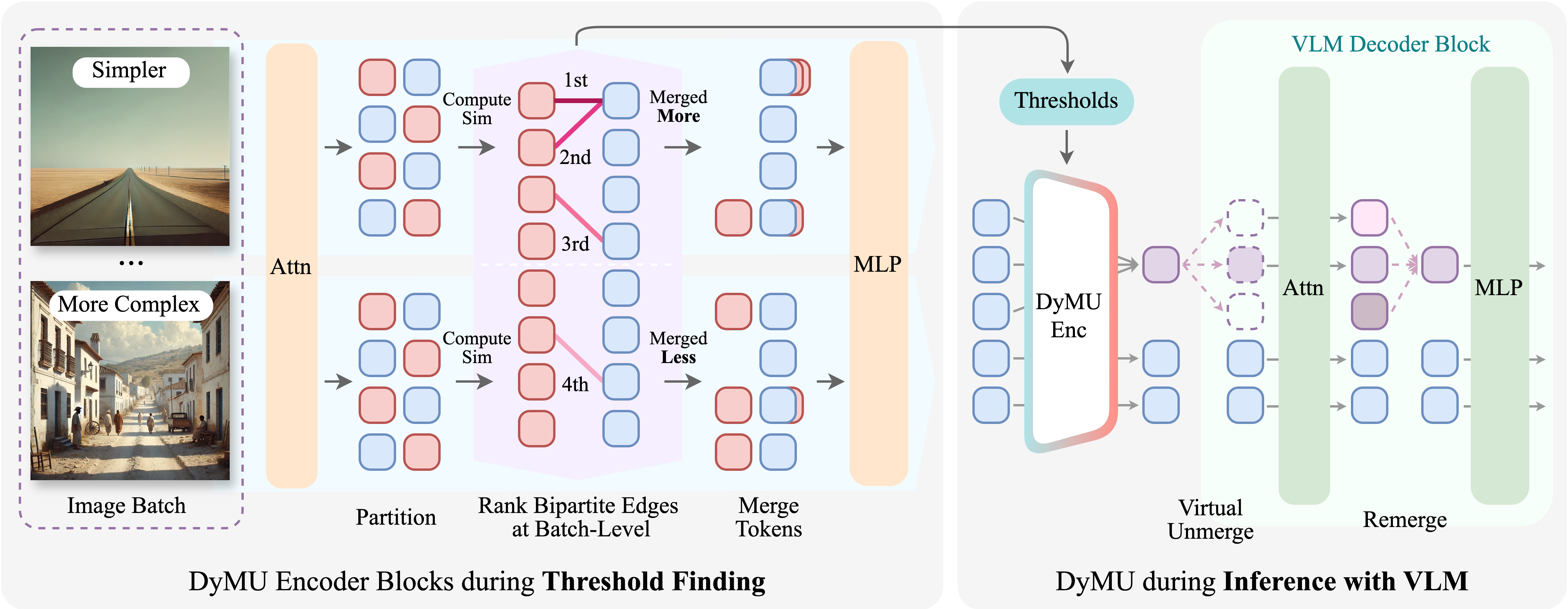}
    \caption{\textbf{Method Overview.} \ours, is composed of two key ideas: Dynamic Token Merging (DToMe) and Virtual Token Unmerging (VTU). 
    DToMe first determines per‐layer thresholds (\textbf{left}) by feeding a large batch of images into the vision transformer and computing bipartite token similarities. We rank these edges across the \textit{entire batch} and choose the top-$Br$ ($r=$ desired \textit{average} number of tokens, batch size $B$). This leads to more edges from simpler images (with more redundancy) being chosen, while complex images remain less merged. During inference, DToMe merges tokens on a per‐image basis using these pre-computed thresholds. We then apply VTU (\textbf{right}) in the self‐attention layers of the pretrained VLM to efficiently expand the attention matrices to the standard token count—ensuring the model’s original weights and outputs remain compatible—before re‐merging the tokens for the next layer. The overall process is training‐free and utilizes crucial image information by allocating the token budget more effectively for both simple and complex images.
    }
    \label{fig:method}
\end{figure*}

%% file: tables/flops.tex
\begin{table}[t]
  \centering
  {\small
  \begin{tabular}{lcc}
    \toprule
    \textbf{Methods} & {$\mathbf{N_{un}/ N}$} & \textbf{MFLOPs} \\ 
    \midrule
    Full Attention & 576 / 576 & 1359.0 \\
    \midrule
    VTU Attention-low & 89 / 576 &  64.9 \\
    VTU Attention-mid & 195 / 576 & 311.5 \\
    VTU Attention-high & 394 / 576 & 1272.0 \\
    \bottomrule
  \end{tabular}
  }
  \caption{Comparison of million floating-point operations per second (MFLOPs) between original attention and Virtual Token Unmerging (VTU) attention. $N$ refers to full sequence length, $N_{un}$ refers to unique sequence length after merging. The statistics are computed with batch size 1, head number 32, and head dimension 128. We use the fvcore package for counting FLOPs.}
  \label{tab:flops}
\end{table}

%% file: tables/main_result_sota.tex
\begin{table*}[ht]
    \centering
    \resizebox{0.98\textwidth}{!}{%
    \begin{minipage}{\textwidth}

    \scriptsize %
    \begin{tabular*}{\textwidth}{@{\extracolsep{\fill}}lcc|ccccccccc|c}

            \toprule
            \textbf{Methods} & \textbf{\begin{tabular}[c]{@{}c@{}}\# Visual \\ Tokens \end{tabular}} & \textbf{\begin{tabular}[c]{@{}c@{}}Compression \\ in Encoder \end{tabular}} & \textbf{GQA} & \textbf{MMB} & \begin{tabular}[c]{@{}c@{}}\textbf{MME} \\ (prcp, all) \end{tabular}
            & \textbf{POPE} & \textbf{SQA$^{I}$} & \textbf{SEED$^{I}$} & \textbf{VQA$^{T}$} & \textbf{MMVet} & \textbf{LLaVA$^W$} & \textbf{Avg} \\
            \midrule
            LLaVA-1.5-7B & 576 & - & 62.0 & 64.6 & 1506,1862 & 86.9 & 69.4 & 66.2 & 58.3 & 30.7 & 63.5 & 63.1 \\
            \midrule
            \rowcolor{gray!10} \multicolumn{13}{c}{{\textit{Fixed Length Compression \& Training-Required }}} \\
            \midrule
            MQT-LLaVA~\cite{MQT-LLAVA} & 256 & No & 61.6 & 64.3 & 1435, - & 84.4 & 67.6 & - & - & 29.8 & 64.6 \\
            Prumerge~\cite{llava-prumerge} & 32 & No & - & 60.9 & 1350, - & 76.3 & 68.5 & - & 56.0 & - & -\\
            Prumerge++~\cite{llava-prumerge} & 144 & No & - & 64.9 & 1462, - & 84.0 & 68.3 & - & 57.1 & - & - \\
            LLaMA-VID~\cite{llama-vid} & 2 & No & 55.5 & - & - , - & 83.1 & 68.8 & - & 49.0 & - & -  \\
            VoCo-LLaMA~\cite{li2024tokenpacker} & 1 & No & 57.0 & 58.8 & 1323, - & 81.4 & 65.4 & - & - & - & - \\
            TokenPacker~\cite{li2024tokenpacker} & 36 & No & 59.6 & 62.8 & - , - & 86.2 & - & - & - & 29.6 & - \\
            LLaVA-Mini~\cite{llava_mini} & 1 & No & 60.9 & 65.6 & 1466, - & 84.4 & 70.4 & - & 57.0 & 36.6 & 68.9 \\
            \midrule
            \rowcolor{gray!10} \multicolumn{13}{c}{{\textit{Fixed Length Compression \& Training-Free}}} \\
            Prumerge-no-ft~\cite{llava-prumerge} & 32 & No & - & - & 1250, - & 76.2 & 68.0 & - & 54.0 & - & - \\
            FastV~\cite{xing2024pyramiddrop} & 128 & No & 49.6 & 56.1 & - , 1490 & 53.4 & 64.4 & - & 50.6 & 26.3 & - \\
            PDrop~\cite{xing2024pyramiddrop} & 128 & No & 56.6 & 61.4 & - , 1713 &  82.3 & 69.2 & - & 55.9 & 30.8 & - \\
            SparseVLM~\cite{zhang2024sparsevlm} & 128 & No & 57.2 & 62.3 & - , 1721 & 85.0 & 67.8 & -  & 55.8 & 29.0 & - \\
            \midrule
            ToMe~\cite{tome} & 94 & {Yes} & 57.3 & 59.7 & 1357, 1673 & 86.8 & 68.9 & 60.5 & 53.2 & 25.6 & 61.0 & 59.2 \\
            ToMe~\cite{tome} & 209 & {Yes} & 59.2 & 62.4 & 1418, 1734 & 87.4 & 69.2 & 63.5 & 54.9 & 30.9 & 62.9 & 61.4 \\
            ToMe~\cite{tome} & 393 & {Yes} & 59.5 & 64.1 & 1454, 1769 & 86.7 & 68.4 & 65.1 & 55.8 & 30.8 & 66.0 & 62.2 \\
            \midrule
            \rowcolor{gray!10} \multicolumn{13}{c}{\textbf{\textit{Variable Length Compression \& Training-Free}}} \\
            \textbf{\ours-low} & $89_{\pm{27}}$ & {Yes} & 60.8 & 62.1 & 1438, 1787 & 86.3 & 69.3 & 65.0 & 53.1 & 30.0 & 62.9 & 61.5 \\
            \textbf{\ours-mid} & $195_{\pm{47}}$ & {Yes} & 61.7 & 62.8 & 1483, 1862 & 86.6 & 69.2 & 65.9 & 55.1 & 30.9 & 65.1 & 62.6 \\
            \textbf{\ours-high} & $394_{\pm{57}}$ & {Yes} & 61.9 & 64.3 & 1498, 1846 & 86.8 & 69.9 & 66.1 & 58.0 & 31.5 & 64.5 & 63.2 \\
            \bottomrule
        \end{tabular*}
    \caption{Comparison with state-of-the-art methods for improving efficiency on LLaVA 1.5~\cite{llava15}. \ours-low achieves 97.5\% of the original full-length LLaVA baseline’s performance while using only $\sim$15\% of the tokens. Importantly, \ours is entirely training-free and generally outperforms previous fixed-length, training-free methods such as \cite{tome, fast-v, zhang2024sparsevlm}, while also enabling variable-length outputs.}
    \label{tab:sota}
    \end{minipage}
    }
\end{table*}

%% file: tables/siglip_results.tex
\begin{table*}[ht]
    \centering
    \setlength{\tabcolsep}{2pt}
    {\scriptsize
    \vspace{-5pt}
\resizebox{0.7\textwidth}{!}
{ %
    \begin{tabular}{@{}l c | c c c c c c c c c | c @{}}
        \toprule
        \textbf{Methods} & \textbf{\begin{tabular}[c]{@{}c@{}}\# Visual \\ Tokens \end{tabular}} & \textbf{GQA} & \textbf{MMB} & \begin{tabular}[c]{@{}c@{}} \textbf{MME}\\(prcp, all)\end{tabular}
        & \textbf{POPE} & \textbf{SQA$^{I}$} & \textbf{SEED$^{I}$} & \textbf{VQA$^{T}$} & \textbf{MMVet} & \textbf{LLaVA$^W$} & \textbf{Avg} \\
        \midrule
        LLaVA-1.5-w-SigLIP & 576 & 62.7 & 65.1 & 1471, 1770 & 85.7 & 68.2 & 66.7 & 57.6 & 30.2 & 59.8 & 62.1 \\
        \midrule
        ToMe~\cite{tome} & 114 & 59.3 & 61.4 & 1380, 1717 & 85.1 & 66.9 & 61.8 & 52.1 & 26.1 & 57.9 & 59.1 \\
        \midrule
        \textbf{\ours-SigLIP-low} & 90$_{\pm{26}}$ & 61.3 & 62.5 & 1398, 1695 & 84.9 & 66.7 & 64.4 & 51.8 & 26.7 & 58.6 & 59.7 \\
        \textbf{\ours-SigLIP-mid} &  176$_{\pm{43}}$ & 62.2 & 63.9 & 1442, 1744 & 85.0 & 67.4 & 65.2 & 54.5 & 26.7 & 59.5 & 60.7 \\
        \textbf{\ours-SigLIP-high} &  318$_{\pm{57}}$ & 62.4 & 65.0 & 1449, 1765 & 86.0 & 67.6 & 66.0 & 56.8 & 29.4 & 58.3 & 61.6 \\
        \bottomrule
    \end{tabular}
    }
    \caption{\ours demonstrates similar efficacy on a different visual encoder, SigLIP~\cite{siglip}.  We obtain the baseline by following the same training recipe as LLaVA-1.5\cite{llava15}. \ours-SigLIP-low achieves $96.1\%$ of the baseline performance while using $\sim$15\% visual tokens.}
    \label{tab:siglip_results}
    }
\end{table*}

%% file: arxiv_sections/3_experiments.tex
\section{Experiments}
\label{sec:experiments}

In this section, we present all the details of our implementation of the proposed method. We also present a comprehensive analysis demonstrating the practical benefits and efficacy of utilizing \ours with various VLMs, visual encoders and LLM architectures.

\subsection{Implementation Details}
\label{sec:implementation}
\paragraph{\dtomefull{}} For \dtome, we find layer-wise thresholds using a diverse dataset of 250k images sampled from the SFT instruction tuning data of LLaVA 1.5~\cite{llava15} comprising of images from MS-COCO~\cite{mscoco}, VisualGenome~\cite{visual_genome}, OCR-VQA~\cite{ocrvqa}, TextVQA~\cite{textvqa} and GQA~\cite{hudson2019gqa}. We also ablate the choice of image datasets in \S\ref{sec:quant}. In general, a sufficiently diverse image set suffices, and performance remains robust to dataset changes. 
Importantly, we only use the images to estimate the thresholds (in inference mode) and do not use the associated annotations or text in any way.

\paragraph{\ours variants} 
For each visual encoder in the experiments, including CLIP~\cite{clip}\footnote{CLIP version: openai/clip-vit-large-patch14-336} and SigLIP~\cite{siglip, siglipso400m}\footnote{SigLIP with LLaVA-1.5: timm/ViT-B-16-SigLIP-384}\footnote{SIgLIP version with LLaVA-OV: google/siglip-so400m-patch14-384}, we find thresholds for three variants of the encoder by choosing different \textit{average} number of tokens to drop ($r_i$) in each layer. We represent these variants by \textit{$\bullet$-low,$\bullet$-mid,$\bullet$-high} corresponding to the expected average number of tokens. 
We also explore different VLM backbones including fixed-resolution models, e.g., LLaVA 1.5~\cite{llava15} and any-resolution models, e.g., LLaVA-OneVision~\cite{llava-ov}.

\subsection{Quantitative Evaluation}
\label{sec:quant}
\paragraph{Comparing Visual Token Merging Methods for VLMs}

In order to evaluate efficacy of our approach, we compare against several existing methods that focus on reducing the number of tokens for VLMs. To the best of our knowledge, our proposed approach is the first to 1) enable varied number of visual tokens and 2) not require further fine-tuning of the VLM. Nevertheless, we compare to methods that are designed to reduce the number of tokens by a fixed length. In Table~\ref{tab:sota}, we present a quantitative evaluation of all methods applied to a pre-trained LLaVA 1.5~\cite{llava15} architecture on standard VLM benchmarks, including GQA~\cite{hudson2019gqa}, MMBench~\cite{liu2023mmbench}, MME~\cite{fu2023mme}, POPE~\cite{pope}, ScienceQA~\cite{scienceqa}, SEED-IMG~\cite{seed_bench}, TextVQA~\cite{textvqa}, MMVet~\cite{yu2023mmvet}, LLaVA-Bench~\cite{llava15}.
\ours achieves average performances of 97.5\%, 99.2\%, and 100.2\%, relative to the original pretrained model, while reducing the token number by 84.5\%, 66.1\%, and 31.6\%, respectively. \ours also outperforms previous training-free methods while enabling varied length output per instance.
When decreasing the token number, the largest drop happens in TextVQA, which fits our expectation as understanding visual text is highly sensitive to the spatial location of visual tokens, on which the token merging tend to break.

\input{tables/llava_ov_results}

\input{figures/complexity}

\input{figures/radar_vtu}

\input{figures/pixmo_vs_llava}

\input{figures/qualitative}

\paragraph{Compatibility with Different LLMs and Visual Encoders}
\ours can be seamlessly integrated into multiple variants of VLMs featuring different LLMs, visual encoders, and pretraining strategies. In Tables~\ref{tab:sota} and \ref{tab:siglip_results}, we demonstrate that \ours effectively maintains baseline performance when applied both CLIP~\cite{clip} to SigLIP~\cite{siglip} representations within the LLaVA 1.5 framework, using a Vicuna-7B~\cite{vicuna2023} LLM.

Furthermore, in Table~\ref{tab:llava_ov_results} we evaluate \ours on LLaVA-OneVision~\cite{llava-ov}, a recent Any-Resolution (AnyRes) model with SigLIP-so400M~\cite{siglipso400m} as visual encoder and Qwen2~\cite{yang2024qwen2technicalreport} as LLM backbone. AnyRes enables processing images of arbitrary resolutions by segmenting them into smaller regions and encoding each individually. Our results show that \ours remains compatible with this complex operation, preserving performance while dynamically reducing token counts.
Additionally, we extend our evaluation to video benchmarks using LLaVA-OneVision. 
By applying \ours to the visual encoder, we achieve a variable reduction in feature representations per frame while maintaining strong performance across benchmarks.

\paragraph{Image Complexity vs Number of Tokens} 
In Figure~\ref{fig:complexity} (left), we show how the number of tokens varies with image complexity. We quantify image complexity $C(I)$ by computing the JPEG compression ratio, i.e., $C(I) = \frac{S_{JPEG}(I)}{H\times W}$, where $S_{JPEG}$ is the size (in bytes) of the image $I$ after JPEG encoding, and $H,W$ are the original height and width.
For this experiment, we use CLIP-L/14-336 with \dtome\textit{-low} to encode images in the MME benchmark. We observe a strong correlation between the number of output tokens and image complexity, indicating that \dtome effectively preserves essential details in complex images while reducing redundancy in simpler ones. We include more qualitative visualizations in Appendix~\ref{app:vis_img_to_tok}.

\paragraph{Fixed vs Dynamic Token Reduction} In Figure~\ref{fig:complexity} (right), we categorize images into three bins based on their complexity scores, and compare the performance of ToMe (fixed-length token reduction) and \dtome on the MME benchmark. A key drawback of fixed token reduction is its inability to adapt to image complexity, leading to over-compression for complex images and under-compression for simpler ones. While our method outperforms ToMe across all complexity levels, we observe the most significant gains on complex images, where ToMe struggles due to an insufficient number of tokens.

\paragraph{Importance of \vtufull}
\vtu efficiently reconstructs the representation of a full visual token sequence from a reduced set of visual tokens
To demonstrate its impact, we compare LLaVA 1.5 variants with and without \vtu. 
In the latter, the LLM does not undergo any modifications and directly receives fewer tokens.
In Figure~\ref{fig:radar_vtu}, we evaluate this effect on two token reduction methods: ToMe~\cite{tome}, which produces fixed-length sequences, and \dtome (ours). Across both cases, we observe that applying \vtu significantly improves performance on 8 out of 9 benchmarks, demonstrating its effectiveness in preserving model capabilities despite token reduction.

\paragraph{Impact of Dataset for Threshold Finding} The \dtome thresholds  are computed using images from the LLaVA instruction tuning dataset. Here, we investigate the sensitivity of \dtome to the threshold estimation dataset. In Figure~\ref{fig:threshold_dataset}, we evaluate \ours-LLaVA 1.5  with \dtome thresholds estimated on the Pixmo-Cap~\cite{deitke2024molmo} image-captioning dataset. We observe a minimal performance change across all the benchmarks, highlighting the robustness of our method to dataset variation. 
Interestingly, we observe that the thresholds estimated using the Pixmo-Cap dataset lead to fewer tokens during inference on the benchmarks. We hypothesize that this is due to the domain shift between the Pixmo-Cap images and a more diverse LLaVA-instruct dataset which covers diverse real-world use cases.

\subsection{Qualitative Analysis}
\label{sec:qualitative}

\vspace{-1pt}
\paragraph{Visualizing Variable Visual Token Length} \dtome facilitates producing variable number of token embeddings for images based on complexity of the content. In Appendix Figure~\ref{fig:img_to_tok_grid}, we visualize the number of visual tokens for various images from nine benchmarks. For each benchmark, we present three images corresponding to the minimum, median, and maximum token numbers output by \ours-low. We observe a strong correlation, both within and across different benchmarks, between image complexity and the number of tokens retained by \ours.

\paragraph{Controllable Visual Token Length}
Dynamic Token Merging offers a key advantage over fixed token reduction methods: cost controllability. By dynamically adjusting the number of visual tokens based on image complexity, users gain direct control over the computational cost incurred per image. 
This flexibility allows flexible combination of visual reasoning tools with \ours to further boost efficiency while maintaining performance.
For instance, in Figure~\ref{fig:qualitative}, we show example applications of combining \ours with additional tools, i.e., background removal~\cite{RMBG14}, OCR~\cite{onnxtr2024}, and object detection~\cite{owlvit} models, to extract focused regions and further reduce token count.
Unlike existing VLMs, which impose a fixed token budget per image regardless of content, our method enables adaptive token allocation, ensuring that simpler regions consume fewer resources while more complex regions retain the necessary level of detail.

%% file: tables/llava_ov_results.tex
\begin{table}[t]
    \centering
    \resizebox{0.48\textwidth}{!}
    { %
    \setlength{\tabcolsep}{2pt}
    \begin{tabular}{@{}lc|cccc|cc}
      \toprule
      \multirow{2}{*}{\textbf{Methods}} & 
      \multirow{2}{*}{\textbf{\begin{tabular}[c]{@{}c@{}}\% Visual\\Tokens\end{tabular}}} 
      & \multicolumn{4}{c|}{\textbf{Image Benchmarks}} 
      & \multicolumn{2}{c}{\textbf{Video Benchmarks}} \\
      &  & \textbf{MMB} & \textbf{MME} 
      & \textbf{SEED} & \textbf{MathVista} & \textbf{VidMME} & \textbf{MMBVid} \\
      \midrule
      LLaVA-ov-7B & 100\% & 79.3 & 75.8 & 75.6 & 58.0 & 61.3 & 1.18 \\
      \midrule
      ToMe~\cite{tome} & 14.4\% & 71.2 & 63.1 & 68.3 & 46.6 & 57.6 & 1.08 \\
      \midrule
      \textbf{\ours-ov-low} & $\sim$14.4\% & 73.6 & 68.0 & 72.9 & 47.4 & 59.3 & 1.08 \\
      \textbf{\ours-ov-mid} & $\sim$25.1\% & 76.0 & 70.3 & 73.7 & 51.7 & 60.1 & 1.12 \\
      \textbf{\ours-ov-high} & $\sim$46.5\% & 77.8 & 73.6 & 74.2 & 54.4 & 60.1 & 1.16 \\
      \bottomrule
    \end{tabular}
    }
    \captionof{table}{\ours shows consistent effectiveness on an AnyRes VLM backbone, 
    LLaVA-OneVision~\cite{llava-ov}. We additionally show performance on two comprehensive video understanding benchmarks, where \ours-ov-low achieves $\sim$96.5\% of the baseline’s performance with only $\sim$14\% tokens.
    }
    \label{tab:llava_ov_results}
\end{table}

%% file: figures/complexity.tex
\begin{figure}[t]
    \centering
    \includegraphics[width=0.9\linewidth]{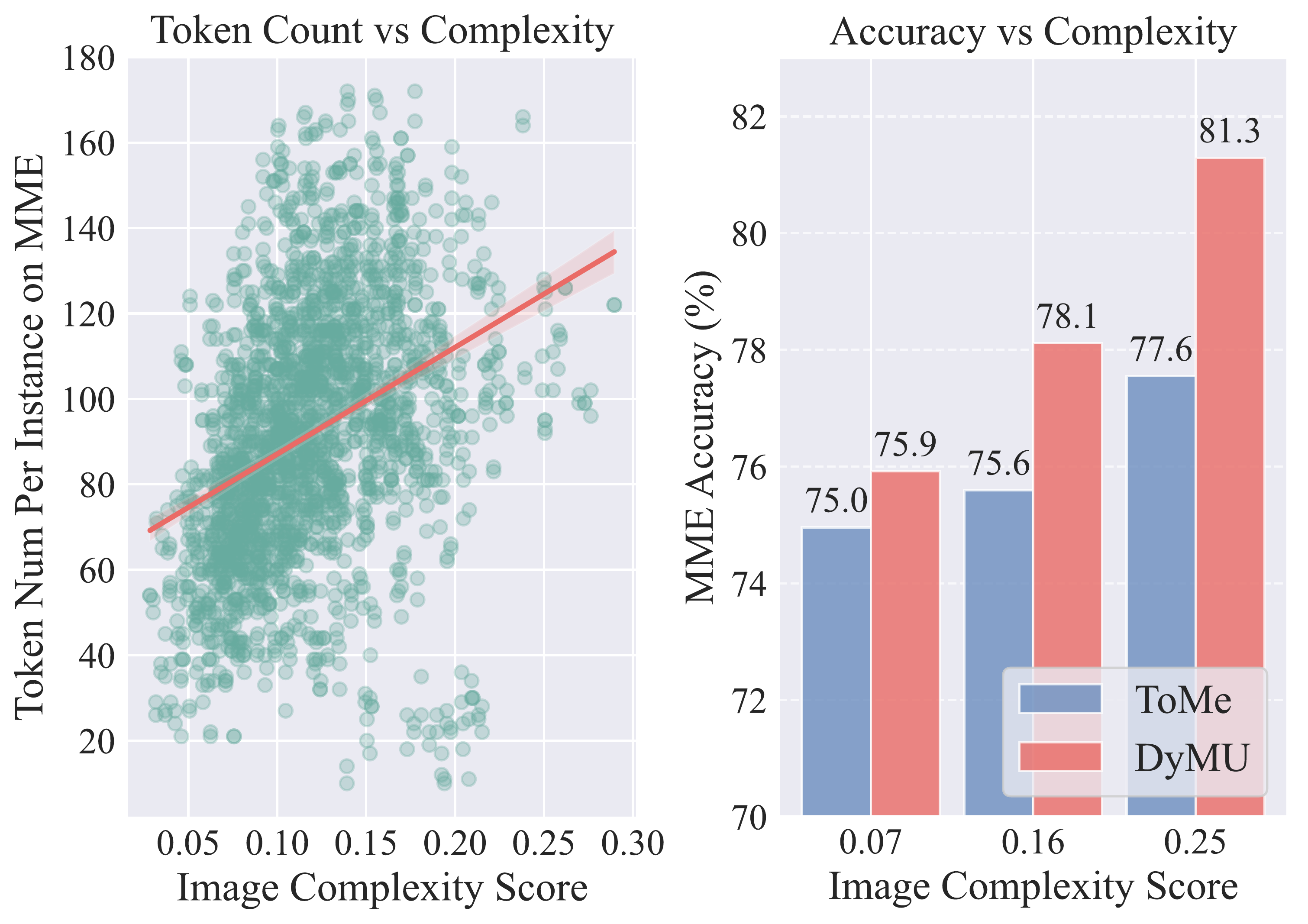}
    \vspace{-3pt}
    \caption{\textbf{Image Complexity vs Token Count and Accuracy} ~ The scatter plot (\textbf{left}) demonstrates a strong correlation between DyMU’s token count and image complexity score—more complex images naturally receive more tokens. On the \textbf{right}, MME accuracy at varying complexity levels is compared between ToMe (fixed-length) and DyMU (dynamic-length), highlighting the benefit of assigning additional tokens to complex images.}
    \label{fig:complexity}
\end{figure}

%% file: figures/radar_vtu.tex
\begin{figure}[t]
    \centering
    \includegraphics[width=0.94\linewidth]{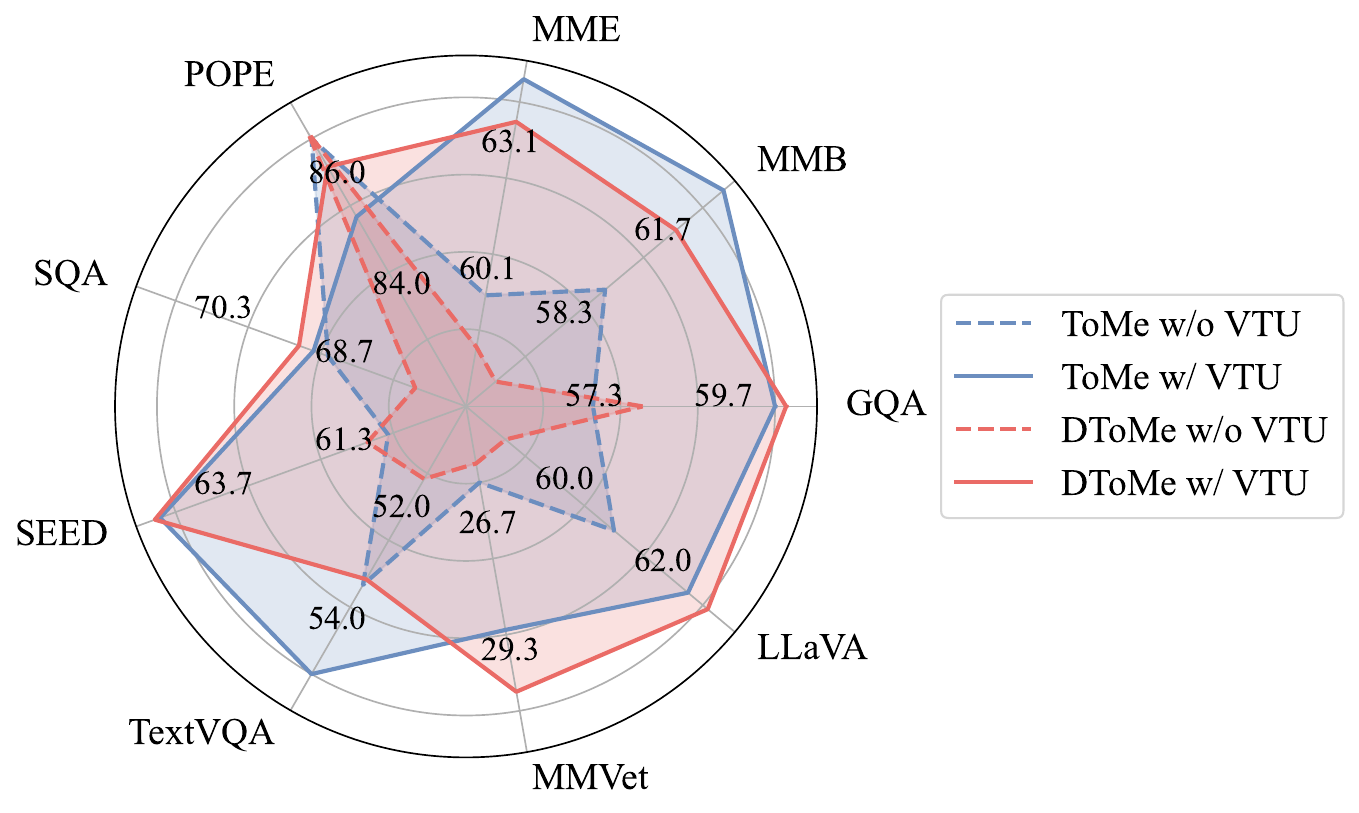}
    \vspace{-5pt}
    \caption{\textbf{Importance of \vtufull(\vtu).} We ablate the performance of LLaVA~1.5 with two token reduction methods applied to the visual encoder—ToMe (fixed‐length) and \dtome (variable‐length). We observe that applying \vtu significantly improves performance on 8 out of 9 benchmarks, demonstrating robustness to varied token reduction methods.}
    \label{fig:radar_vtu}
\end{figure}

%% file: figures/pixmo_vs_llava.tex
\begin{figure}[t]
    \centering
    \includegraphics[width=0.82\linewidth]{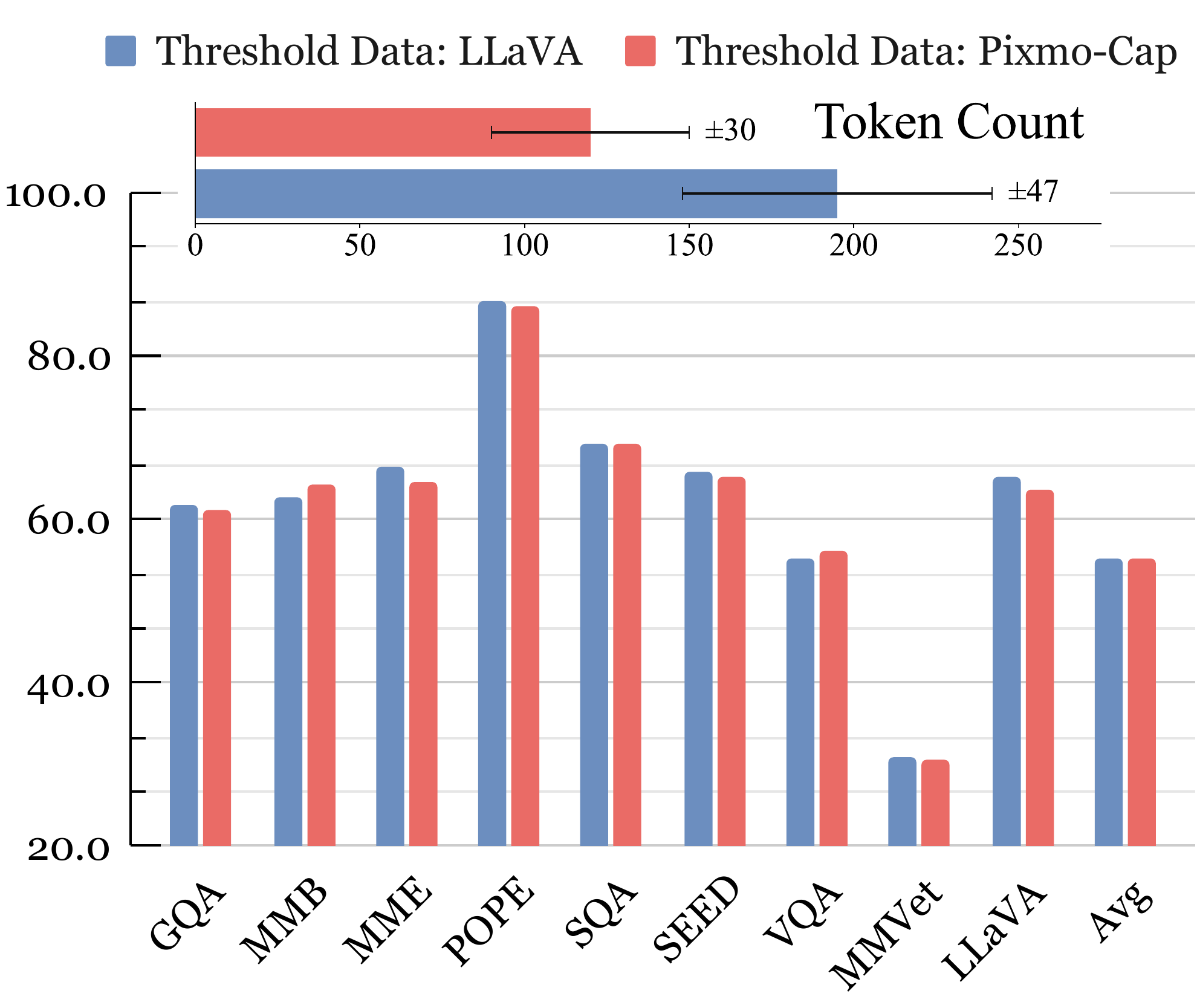}
    \vspace{-3pt}
    \caption{\textbf{Comparing thresholds using LLaVA Instruct Data vs Pixmo-Cap}. Although both methods use the same per‐layer merging hyperparameter ($r_i$ ), the Pixmo‐based thresholds lead to fewer tokens (\textbf{top})—likely due to domain differences.
    However, performance across a range of benchmarks shows minimal drop (\textbf{bottom}),
    indicating the robustness of our threshold estimation.}
    \label{fig:threshold_dataset}
\end{figure}

%% file: figures/qualitative.tex
\begin{figure*}[ht]
    \centering
    \includegraphics[width=0.97\linewidth]{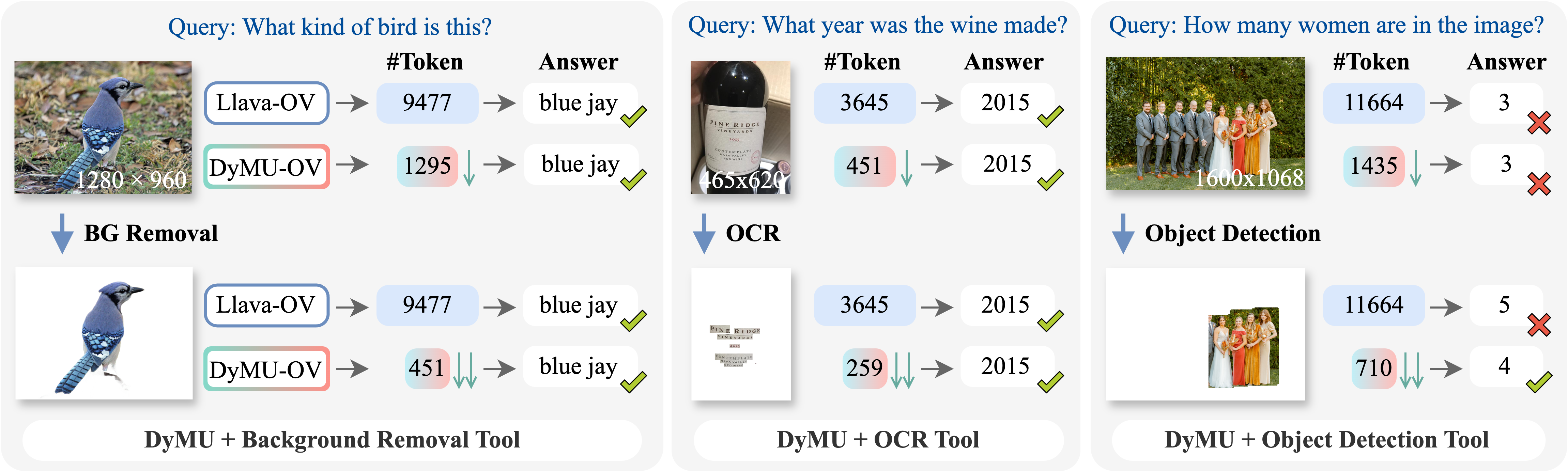}
    \vspace{-1pt}
    \caption{\textbf{Controllable Visual Token Length.} By dynamically allocating tokens based on image complexity, \ours enables direct control over computational cost. In these examples, we combine \ours with additional vision tools—background removal, OCR, or object detection—to focus only on the relevant regions. As a result, token count is substantially reduced without degrading performance, showcasing the flexibility of \ours to adapt token usage according to the task’s requirements.}
    \label{fig:qualitative}
\end{figure*}

%% file: arxiv_sections/5_conclusion.tex
\section{Conclusions and Future Work}
\vspace{-1pt}
In this work, we introduced \ours, the first training-free framework that dynamically reduces visual token counts in VLMs based on per-image complexity. \ours can be directly plugged into all mainstream VLMs that comprise ViT-based visual encoders and RoPE-based LLM backbones. Future work includes improving \ours's ability to preserve VLM performance on spatially sensitive tasks such as TextVQA~\cite{textvqa} and spatial reasoning. Additionally, exploring the extension of \ours to reduce temporal redundancy in videos is another promising direction.

%% file: arxiv_sections/appendix.tex
\appendix

\section{Visualization of Variable Token Length}
\label{app:vis_img_to_tok}

In Figure~\ref{fig:img_to_tok_grid}, we present a comprehensive visualization of example images along with their encoded visual token counts. We use \ours-low (based on CLIP-L/14-336) as the encoder, where the full token length is 576. Three images are shown for each benchmark, corresponding to the minimum, median, and maximum number of tokens, respectively. A clear correlation can be observed between semantic richness and token count. We also note variations in the token range across different benchmarks. For instance, ScienceQA~\cite{scienceqa}, which primarily contains figures and charts, tends to have fewer tokens than benchmarks featuring complex real-world scenes.

\input{figures/img_to_tok_grid}

\section{Impact of Token Merging Schedule}
\input{tables/scheduling}

We conduct an additional ablation study on one of the hyperparameters in \dtome, the merging schedule, during threshold finding. As detailed in Section~\ref{sec:method}, we set a target reduction number,  $r_i$ , for each layer. By default,  $r_i$  is set to a constant value across all layers. Alternatively, we can vary  $r_i$  across layers to encourage merging more or fewer tokens at different depths.

In Table~\ref{tab:scheduling}, we present an ablation study on two alternative scheduling strategies: (1) \textit{linear}, which merges \textit{more} tokens in \textit{earlier} layers and \textit{fewer} tokens in \textit{later} layers, and (2) \textit{reverse linear}, which follows the opposite trend. 
The results indicate that merging fewer tokens in earlier layers tends to yield better performance, while the constant schedule provides a balanced trade-off between performance and token count. This observation echoes the findings in the ToMe paper~\cite{tome}, where a constant schedule was found to be nearly optimal.

\section{Full Results for Figure~\ref{fig:radar_vtu}}
\input{tables/virtual_expansion}
We present the complete results of the ablation experiments on the effect of our proposed \vtufull, as shown in Figure~\ref{fig:radar_vtu}. The results are provided in Table~\ref{tab:vtu}.

\section{Full Results for Figure~\ref{fig:threshold_dataset}}
\input{tables/pixmo}
We present the complete results of the ablation experiments on threshold-finding datasets, as shown in Figure~\ref{fig:threshold_dataset}. The results are provided in Table~\ref{tab:pixmo}.

%% file: figures/img_to_tok_grid.tex
\begin{figure*}[ht]
    \centering
    \includegraphics[width=0.99\linewidth]{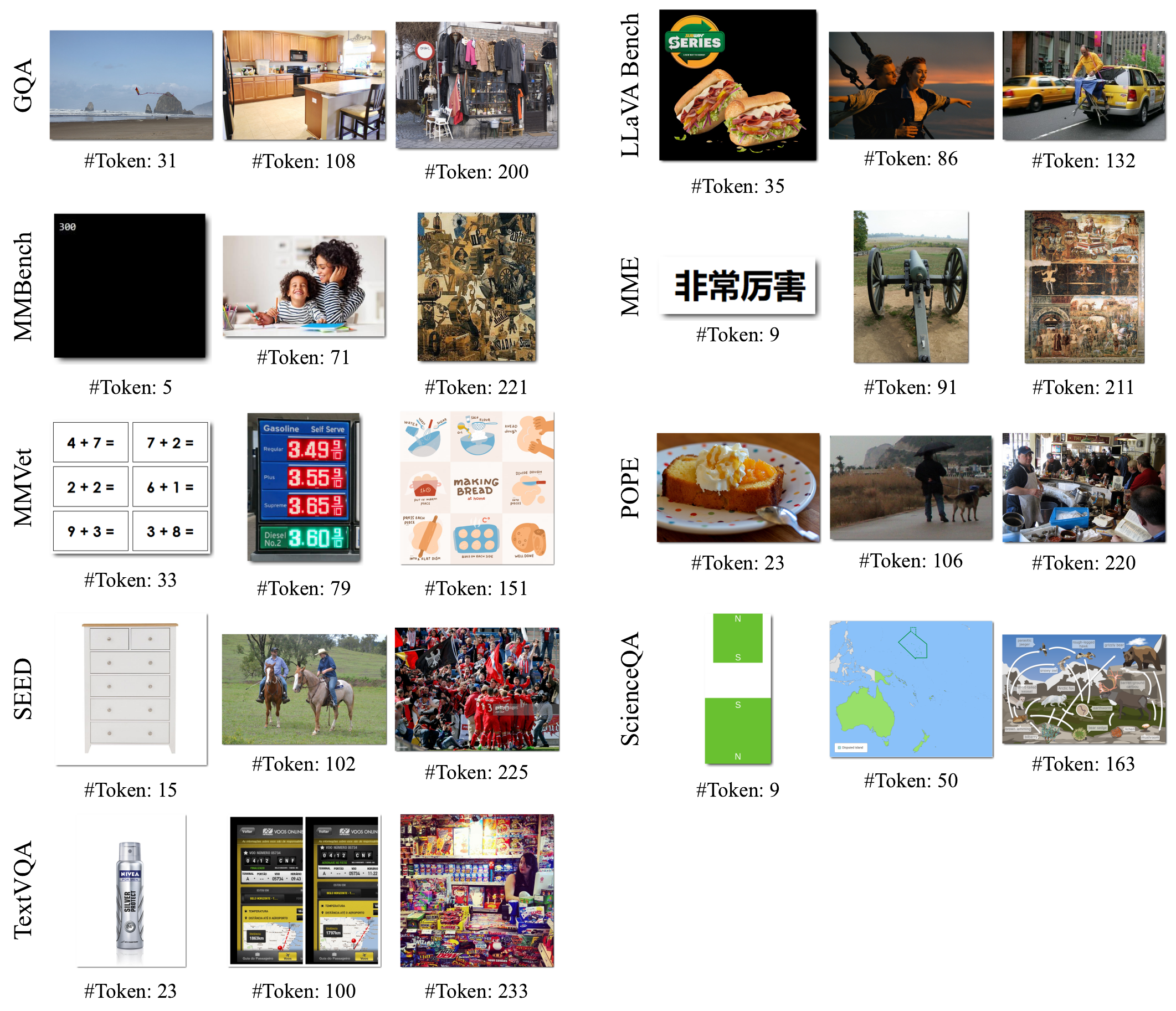}
    \caption{\textbf{\dtome Token Count Across Benchmarks.} For each dataset, we show three examples processed by our method—those yielding the fewest tokens, the median number of tokens, and the most tokens. Observe that visually simple or nearly blank images consistently require fewer tokens, while more detailed, semantically complex or cluttered images produce more tokens. This demonstrates how \dtome effectively adapts to image complexity across diverse benchmarks, allocating fewer tokens to simpler content and preserving more tokens for complex scenes.}
    \label{fig:img_to_tok_grid}
\end{figure*}

%% file: tables/scheduling.tex
\begin{table*}[htb]
    \centering
    \resizebox{0.98\textwidth}{!}
    { %
    \begin{tabular}{@{}l c | c c c c c c c c c | c @{}}
        \toprule
        \textbf{Schedule} & \textbf{\begin{tabular}[c]{@{}c@{}}\# Visual \\ Tokens \end{tabular}} & \textbf{GQA} & \textbf{MMB} & \begin{tabular}[c]{@{}c@{}}\textbf{MME$^{(prcp, all)}$}\end{tabular}
        & \textbf{POPE} & \textbf{SQA$^{I}$} & \textbf{SEED$^{I}$} & \textbf{VQA$^{T}$} & \textbf{MMVet} & \textbf{LLaVA$^W$} & \textbf{Avg} \\
        \midrule
        Constant & 195$_{\pm{47}}$ & 61.7 & 62.8 & 1483, 1862 & 86.6 & 69.2 & 65.9 & 55.1 & 30.9 & 65.1 & 62.6 \\
        Linear &  163$_{\pm{43}}$ & 61.3 & 62.3 & 1437, 1767 & 86.2 & 69.4 & 65.3 & 52.1 & 28.8 & 58.6 & 60.8 \\
        Reverse Linear & 213$_{\pm{49}}$ & 61.8 & 63.8 & 1491, 1863 & 86.7 & 69.3 & 66.0 & 57.5 & 31.8 & 65.3 & 63.2 \\
        \bottomrule
    \end{tabular}
    }
    \caption{Ablation study on merging schedules in \dtome. We compare three strategies: constant, linear (more merging in early layers), and reverse linear (more merging in later layers). Results show that merging fewer tokens in early layers yields better performance, while the constant schedule provides a balanced trade-off between performance and token count. }
    \label{tab:scheduling}
\end{table*}

%% file: tables/virtual_expansion.tex
\begin{table*}[ht]
    \centering
    \small
    \resizebox{0.98\textwidth}{!}
    { %
    \begin{tabular}{@{}l c | c c c c c c c c c | c @{}}
        \toprule
        \textbf{Method} & \textbf{\begin{tabular}[c]{@{}c@{}}\# Visual \\ Tokens \end{tabular}} & \textbf{GQA} & \textbf{MMB} & \begin{tabular}[c]{@{}c@{}}\textbf{MME$^{(prcp, all)}$}\end{tabular}
        & \textbf{POPE} & \textbf{SQA$^{I}$} & \textbf{SEED$^{I}$} & \textbf{VQA$^{T}$} & \textbf{MMVet} & \textbf{LLaVA$^W$} & \textbf{Avg} \\
        \midrule
        ToMe~\cite{tome} & 94 & 57.3 & 59.7 & 1357, 1673 & 86.8 & 68.9 & 60.5 & 53.2 & 25.6 & 61.0 & 59.2  \\
         \quad \textcolor{MyDarkGreen}{\textbf{+ \vtu}} & 94 & 60.6 & 63.7 & 1464, 1815 & 85.4 & 69.1 & 64.9 & 54.8 & 28.7 & 62.5 & 61.6 \\
        \midrule
        \ours-low & $89_{\pm{27}}$ & 60.8 & 62.1 & 1438, 1787 & 86.3 & 69.3 & 65.0 & 53.1 & 30.0 & 62.9 & 61.5 \\
        \quad \textcolor{MyDarkRed}{\textbf{w/o \vtu}} & $89_{\pm{27}}$ & 58.2 & 56.0 & 1346, 1639 & 86.9 & 67.7 & 60.9 & 51.3 & 25.2 & 58.8 & 58.2 \\
        \bottomrule
    \end{tabular}
    }
    \caption{Impact of \vtufull{}. Full results for Figure~\ref{fig:radar_vtu}.}
    \label{tab:vtu}
\end{table*}

%% file: tables/pixmo.tex
\begin{table*}[htb]
    \centering
    {\normalsize
    \resizebox{0.98\textwidth}{!}
    { %
    \begin{tabular}{@{}l c c | c c c c c c c c c | c @{}}
        \toprule
        \textbf{Model} & \textbf{\begin{tabular}[c]{@{}c@{}} Thresh Finding \\ Dataset \end{tabular}} &  \textbf{\begin{tabular}[c]{@{}c@{}}\# Visual \\ Tokens \end{tabular}} & \textbf{GQA} & \textbf{MMB} & \begin{tabular}[c]{@{}c@{}}\textbf{MME$^{(prcp, all)}$}\end{tabular}
        & \textbf{POPE} & \textbf{SQA$^{I}$} & \textbf{SEED$^{I}$} & \textbf{VQA$^{T}$} & \textbf{MMVet} & \textbf{LLaVA$^W$} & \textbf{Avg} \\
        \midrule
        \textbf{\ours-mid} & \textbf{Llava} & 195$_{\pm{47}}$ & 61.7 & 62.8 & 1483, 1862 & 86.6 & 69.2 & 65.9 & 55.1 & 30.9 & 65.1 & 62.6 \\
        \textbf{\ours-mid} & \textbf{ Pixmo} & 120$_{\pm{30}}$ & 61.1 & 64.4 & 1474, 1808 & 86.0 & 69.4 & 65.3 & 56.2 & 30.5 & 63.7 & 62.4 \\
        \bottomrule
    \end{tabular}
    }
    \caption{Impact of dataset for threshold finding. Full results for Figure~\ref{fig:threshold_dataset}.}
    \label{tab:pixmo}
    }
\end{table*}

%% file: main_arxiv.bbl
\begin{thebibliography}{47}
\providecommand{\natexlab}[1]{#1}
\providecommand{\url}[1]{\texttt{#1}}
\expandafter\ifx\csname urlstyle\endcsname\relax
  \providecommand{\doi}[1]{doi: #1}\else
  \providecommand{\doi}{doi: \begingroup \urlstyle{rm}\Url}\fi

\bibitem[Alabdulmohsin et~al.(2023)Alabdulmohsin, Zhai, Kolesnikov, and Beyer]{siglipso400m}
Ibrahim~M Alabdulmohsin, Xiaohua Zhai, Alexander Kolesnikov, and Lucas Beyer.
\newblock Getting vit in shape: Scaling laws for compute-optimal model design.
\newblock \emph{Advances in Neural Information Processing Systems}, 36:\penalty0 16406--16425, 2023.

\bibitem[Bai et~al.(2025)Bai, Chen, Liu, Wang, Ge, Song, Dang, Wang, Wang, Tang, et~al.]{qwen25vl}
Shuai Bai, Keqin Chen, Xuejing Liu, Jialin Wang, Wenbin Ge, Sibo Song, Kai Dang, Peng Wang, Shijie Wang, Jun Tang, et~al.
\newblock Qwen2. 5-vl technical report.
\newblock \emph{arXiv preprint arXiv:2502.13923}, 2025.

\bibitem[Bolya et~al.(2022)Bolya, Fu, Dai, Zhang, Feichtenhofer, and Hoffman]{tome}
Daniel Bolya, Cheng-Yang Fu, Xiaoliang Dai, Peizhao Zhang, Christoph Feichtenhofer, and Judy Hoffman.
\newblock Token merging: Your vit but faster.
\newblock \emph{arXiv preprint arXiv:2210.09461}, 2022.

\bibitem[{Bria AI}(2024)]{RMBG14}
{Bria AI}.
\newblock {RMBG-1.4: Background Removal Model}.
\newblock \url{https://huggingface.co/briaai/RMBG-1.4}, 2024.

\bibitem[Chen et~al.(2024)Chen, Zhao, Liu, Bai, Lin, Zhou, and Chang]{fast-v}
Liang Chen, Haozhe Zhao, Tianyu Liu, Shuai Bai, Junyang Lin, Chang Zhou, and Baobao Chang.
\newblock An image is worth 1/2 tokens after layer 2: Plug-and-play inference acceleration for large vision-language models.
\newblock In \emph{European Conference on Computer Vision}, pages 19--35. Springer, 2024.

\bibitem[Chen et~al.(2015)Chen, Fang, Lin, Vedantam, Gupta, Doll{\'{a}}r, and Zitnick]{mscoco_caption}
Xinlei Chen, Hao Fang, Tsung{-}Yi Lin, Ramakrishna Vedantam, Saurabh Gupta, Piotr Doll{\'{a}}r, and C.~Lawrence Zitnick.
\newblock {Microsoft COCO Captions: Data Collection and Evaluation Server}.
\newblock \emph{CoRR}, abs/1504.00325, 2015.

\bibitem[Chiang et~al.(2023)Chiang, Li, Lin, Sheng, Wu, Zhang, Zheng, Zhuang, Zhuang, Gonzalez, Stoica, and Xing]{vicuna2023}
Wei-Lin Chiang, Zhuohan Li, Zi Lin, Ying Sheng, Zhanghao Wu, Hao Zhang, Lianmin Zheng, Siyuan Zhuang, Yonghao Zhuang, Joseph~E. Gonzalez, Ion Stoica, and Eric~P. Xing.
\newblock {Vicuna: An Open-Source Chatbot Impressing GPT-4 with 90\%* ChatGPT Quality}, 2023.

\bibitem[Deitke et~al.(2024)Deitke, Clark, Lee, Tripathi, Yang, Park, Salehi, Muennighoff, Lo, Soldaini, et~al.]{deitke2024molmo}
Matt Deitke, Christopher Clark, Sangho Lee, Rohun Tripathi, Yue Yang, Jae~Sung Park, Mohammadreza Salehi, Niklas Muennighoff, Kyle Lo, Luca Soldaini, et~al.
\newblock Molmo and pixmo: Open weights and open data for state-of-the-art multimodal models.
\newblock \emph{arXiv preprint arXiv:2409.17146}, 2024.

\bibitem[Dittrich(2024)]{onnxtr2024}
Felix Dittrich.
\newblock Onnxtr: Optical character recognition made seamless \& accessible to anyone, powered by onnx.
\newblock \url{https://github.com/felixdittrich92/OnnxTR}, 2024.

\bibitem[Francis and Kucera(1979)]{francis79browncorpus}
W.~N. Francis and H. Kucera.
\newblock Brown corpus manual.
\newblock Technical report, Department of Linguistics, Brown University, Providence, Rhode Island, US, 1979.

\bibitem[Fu et~al.(2023)Fu, Chen, Shen, Qin, Zhang, Lin, Qiu, Lin, Yang, Zheng, Li, Sun, and Ji]{fu2023mme}
Chaoyou Fu, Peixian Chen, Yunhang Shen, Yulei Qin, Mengdan Zhang, Xu Lin, Zhenyu Qiu, Wei Lin, Jinrui Yang, Xiawu Zheng, Ke Li, Xing Sun, and Rongrong Ji.
\newblock {{MME:} {A} Comprehensive Evaluation Benchmark for Multimodal Large Language Models}.
\newblock \emph{arXiv prepring arXiv:2306.13394}, 2023.

\bibitem[Goyal et~al.(2017)Goyal, Khot, Summers{-}Stay, Batra, and Parikh]{vqav2}
Yash Goyal, Tejas Khot, Douglas Summers{-}Stay, Dhruv Batra, and Devi Parikh.
\newblock {Making the {V} in {VQA} Matter: Elevating the Role of Image Understanding in Visual Question Answering}.
\newblock In \emph{2017 {IEEE} Conference on Computer Vision and Pattern Recognition (CVPR)}, 2017.

\bibitem[Gupta et~al.(2019)Gupta, Dollar, and Girshick]{gupta2019lvis}
Agrim Gupta, Piotr Dollar, and Ross Girshick.
\newblock {LVIS}: A dataset for large vocabulary instance segmentation.
\newblock In \emph{Proceedings of the {IEEE} Conference on Computer Vision and Pattern Recognition}, 2019.

\bibitem[Hu et~al.(2024)Hu, Dou, Li, Kamath, Peng, and Chang]{MQT-LLAVA}
Wenbo Hu, Zi-Yi Dou, Liunian~Harold Li, Amita Kamath, Nanyun Peng, and Kai-Wei Chang.
\newblock Matryoshka query transformer for large vision-language models.
\newblock \emph{arXiv preprint arXiv:2405.19315}, 2024.

\bibitem[Huang et~al.(2024)Huang, Zhai, Shen, Cao, Zhao, Xu, Ye, and Lin]{dynamic-llava}
Wenxuan Huang, Zijie Zhai, Yunhang Shen, Shaoshen Cao, Fei Zhao, Xiangfeng Xu, Zheyu Ye, and Shaohui Lin.
\newblock Dynamic-llava: Efficient multimodal large language models via dynamic vision-language context sparsification.
\newblock \emph{arXiv preprint arXiv:2412.00876}, 2024.

\bibitem[Hudson and Manning(2019)]{hudson2019gqa}
Drew~A Hudson and Christopher~D Manning.
\newblock Gqa: A new dataset for real-world visual reasoning and compositional question answering.
\newblock In \emph{Proceedings of the IEEE/CVF conference on computer vision and pattern recognition}, pages 6700--6709, 2019.

\bibitem[Krishna et~al.(2017)Krishna, Zhu, Groth, Johnson, Hata, Kravitz, Chen, Kalantidis, Li, Shamma, Bernstein, and Fei{-}Fei]{visual_genome}
Ranjay Krishna, Yuke Zhu, Oliver Groth, Justin Johnson, Kenji Hata, Joshua Kravitz, Stephanie Chen, Yannis Kalantidis, Li{-}Jia Li, David~A. Shamma, Michael~S. Bernstein, and Li Fei{-}Fei.
\newblock {Visual Genome: Connecting Language and Vision Using Crowdsourced Dense Image Annotations}.
\newblock \emph{Int. J. Comput. Vis.}, 123\penalty0 (1):\penalty0 32--73, 2017.

\bibitem[Li et~al.(2023{\natexlab{a}})Li, Wang, Wang, Ge, Ge, and Shan]{seed_bench}
Bohao Li, Rui Wang, Guangzhi Wang, Yuying Ge, Yixiao Ge, and Ying Shan.
\newblock {SEED-Bench: Benchmarking Multimodal LLMs with Generative Comprehension}.
\newblock \emph{arXiv prepring arXiv:2307.16125}, 2023{\natexlab{a}}.

\bibitem[Li et~al.(2024{\natexlab{a}})Li, Zhang, Guo, Zhang, Li, Zhang, Zhang, Li, Liu, and Li]{llava-ov}
Bo Li, Yuanhan Zhang, Dong Guo, Renrui Zhang, Feng Li, Hao Zhang, Kaichen Zhang, Yanwei Li, Ziwei Liu, and Chunyuan Li.
\newblock Llava-onevision: Easy visual task transfer.
\newblock \emph{arXiv preprint arXiv:2408.03326}, 2024{\natexlab{a}}.

\bibitem[Li et~al.(2024{\natexlab{b}})Li, Yuan, Liu, Tang, Wang, Qin, Zhu, and Zhang]{li2024tokenpacker}
Wentong Li, Yuqian Yuan, Jian Liu, Dongqi Tang, Song Wang, Jie Qin, Jianke Zhu, and Lei Zhang.
\newblock Tokenpacker: Efficient visual projector for multimodal llm.
\newblock \emph{arXiv preprint arXiv:2407.02392}, 2024{\natexlab{b}}.

\bibitem[Li et~al.(2023{\natexlab{b}})Li, Du, Zhou, Wang, Zhao, and Wen]{pope}
Yifan Li, Yifan Du, Kun Zhou, Jinpeng Wang, Wayne~Xin Zhao, and Ji-Rong Wen.
\newblock Evaluating object hallucination in large vision-language models.
\newblock \emph{arXiv preprint arXiv:2305.10355}, 2023{\natexlab{b}}.

\bibitem[Li et~al.(2024{\natexlab{c}})Li, Wang, and Jia]{llama-vid}
Yanwei Li, Chengyao Wang, and Jiaya Jia.
\newblock Llama-vid: An image is worth 2 tokens in large language models.
\newblock In \emph{European Conference on Computer Vision}, pages 323--340. Springer, 2024{\natexlab{c}}.

\bibitem[Liang et~al.(2025)Liang, Guan, Lu, Chen, Wang, and Hu]{dynamic_token_reduction}
Xiaoyu Liang, Chaofeng Guan, Jiaying Lu, Huiyao Chen, Huan Wang, and Haoji Hu.
\newblock Dynamic token reduction during generation for vision language models.
\newblock \emph{arXiv preprint arXiv:2501.14204}, 2025.

\bibitem[Lin et~al.(2014)Lin, Maire, Belongie, Hays, Perona, Ramanan, Doll{\'{a}}r, and Zitnick]{mscoco}
Tsung{-}Yi Lin, Michael Maire, Serge~J. Belongie, James Hays, Pietro Perona, Deva Ramanan, Piotr Doll{\'{a}}r, and C.~Lawrence Zitnick.
\newblock {Microsoft COCO: Common Objects in Context}.
\newblock In \emph{Computer Vision - {ECCV} 2014 - 13th European Conference}, 2014.

\bibitem[Liu et~al.(2023{\natexlab{a}})Liu, Li, Li, and Lee]{llava15}
Haotian Liu, Chunyuan Li, Yuheng Li, and Yong~Jae Lee.
\newblock {Improved Baselines with Visual Instruction Tuning}.
\newblock \emph{arXiv prepring arXiv:2310.03744}, 2023{\natexlab{a}}.

\bibitem[Liu et~al.(2022)Liu, Wu, and Guo]{adaptive_sparse_vit}
Xiangcheng Liu, Tianyi Wu, and Guodong Guo.
\newblock Adaptive sparse vit: Towards learnable adaptive token pruning by fully exploiting self-attention.
\newblock \emph{arXiv preprint arXiv:2209.13802}, 2022.

\bibitem[Liu et~al.(2023{\natexlab{b}})Liu, Duan, Zhang, Li, Zhang, Zhao, Yuan, Wang, He, Liu, Chen, and Lin]{liu2023mmbench}
Yuan Liu, Haodong Duan, Yuanhan Zhang, Bo Li, Songyang Zhang, Wangbo Zhao, Yike Yuan, Jiaqi Wang, Conghui He, Ziwei Liu, Kai Chen, and Dahua Lin.
\newblock {MMBench: Is Your Multi-modal Model an All-around Player?}
\newblock \emph{arXiv prepring arXiv:2307.06281}, 2023{\natexlab{b}}.

\bibitem[Lu et~al.(2022)Lu, Mishra, Xia, Qiu, Chang, Zhu, Tafjord, Clark, and Kalyan]{scienceqa}
Pan Lu, Swaroop Mishra, Tony Xia, Liang Qiu, Kai-Wei Chang, Song-Chun Zhu, Oyvind Tafjord, Peter Clark, and Ashwin Kalyan.
\newblock Learn to explain: Multimodal reasoning via thought chains for science question answering.
\newblock In \emph{The 36th Conference on Neural Information Processing Systems (NeurIPS)}, 2022.

\bibitem[Marin et~al.(2021)Marin, Chang, Ranjan, Prabhu, Rastegari, and Tuzel]{token_pooling}
Dmitrii Marin, Jen-Hao~Rick Chang, Anurag Ranjan, Anish Prabhu, Mohammad Rastegari, and Oncel Tuzel.
\newblock Token pooling in vision transformers.
\newblock \emph{arXiv preprint arXiv:2110.03860}, 2021.

\bibitem[Minderer et~al.(2022)Minderer, Gritsenko, Stone, Neumann, Weissenborn, Dosovitskiy, Mahendran, Arnab, Dehghani, Shen, et~al.]{owlvit}
Matthias Minderer, Alexey Gritsenko, Austin Stone, Maxim Neumann, Dirk Weissenborn, Alexey Dosovitskiy, Aravindh Mahendran, Anurag Arnab, Mostafa Dehghani, Zhuoran Shen, et~al.
\newblock Simple open-vocabulary object detection.
\newblock In \emph{European conference on computer vision}, pages 728--755. Springer, 2022.

\bibitem[Mishra et~al.(2019)Mishra, Shekhar, Singh, and Chakraborty]{ocrvqa}
Anand Mishra, Shashank Shekhar, Ajeet~Kumar Singh, and Anirban Chakraborty.
\newblock Ocr-vqa: Visual question answering by reading text in images.
\newblock In \emph{ICDAR}, 2019.

\bibitem[Oquab et~al.(2023)Oquab, Darcet, Moutakanni, Vo, Szafraniec, Khalidov, Fernandez, Haziza, Massa, El{-}Nouby, Assran, Ballas, Galuba, Howes, Huang, Li, Misra, Rabbat, Sharma, Synnaeve, Xu, J{\'{e}}gou, Mairal, Labatut, Joulin, and Bojanowski]{oquab2023dinov2}
Maxime Oquab, Timoth{\'{e}}e Darcet, Th{\'{e}}o Moutakanni, Huy Vo, Marc Szafraniec, Vasil Khalidov, Pierre Fernandez, Daniel Haziza, Francisco Massa, Alaaeldin El{-}Nouby, Mahmoud Assran, Nicolas Ballas, Wojciech Galuba, Russell Howes, Po{-}Yao Huang, Shang{-}Wen Li, Ishan Misra, Michael~G. Rabbat, Vasu Sharma, Gabriel Synnaeve, Hu Xu, Herv{\'{e}} J{\'{e}}gou, Julien Mairal, Patrick Labatut, Armand Joulin, and Piotr Bojanowski.
\newblock {DINOv2: Learning Robust Visual Features without Supervision}.
\newblock \emph{arXiv prepring arXiv:2304.07193}, 2023.

\bibitem[Radford et~al.(2021)Radford, Kim, Hallacy, Ramesh, Goh, Agarwal, Sastry, Askell, Mishkin, Clark, Krueger, and Sutskever]{clip}
Alec Radford, Jong~Wook Kim, Chris Hallacy, Aditya Ramesh, Gabriel Goh, Sandhini Agarwal, Girish Sastry, Amanda Askell, Pamela Mishkin, Jack Clark, Gretchen Krueger, and Ilya Sutskever.
\newblock {Learning Transferable Visual Models From Natural Language Supervision}.
\newblock In \emph{Proceedings of the 38th International Conference on Machine Learning (ICML)}, 2021.

\bibitem[Shang et~al.(2024)Shang, Cai, Xu, Lee, and Yan]{llava-prumerge}
Yuzhang Shang, Mu Cai, Bingxin Xu, Yong~Jae Lee, and Yan Yan.
\newblock Llava-prumerge: Adaptive token reduction for efficient large multimodal models.
\newblock \emph{arXiv preprint arXiv:2403.15388}, 2024.

\bibitem[Singh et~al.(2019)Singh, Natarjan, Shah, Jiang, Chen, Parikh, and Rohrbach]{textvqa}
Amanpreet Singh, Vivek Natarjan, Meet Shah, Yu Jiang, Xinlei Chen, Devi Parikh, and Marcus Rohrbach.
\newblock Towards vqa models that can read.
\newblock In \emph{Proceedings of the IEEE Conference on Computer Vision and Pattern Recognition}, pages 8317--8326, 2019.

\bibitem[Su et~al.(2024)Su, Ahmed, Lu, Pan, Bo, and Liu]{rope}
Jianlin Su, Murtadha Ahmed, Yu Lu, Shengfeng Pan, Wen Bo, and Yunfeng Liu.
\newblock Roformer: Enhanced transformer with rotary position embedding.
\newblock \emph{Neurocomputing}, 568:\penalty0 127063, 2024.

\bibitem[Wang et~al.(2024)Wang, Bai, Tan, Wang, Fan, Bai, Chen, Liu, Wang, Ge, Fan, Dang, Du, Ren, Men, Liu, Zhou, Zhou, and Lin]{Qwen2VL}
Peng Wang, Shuai Bai, Sinan Tan, Shijie Wang, Zhihao Fan, Jinze Bai, Keqin Chen, Xuejing Liu, Jialin Wang, Wenbin Ge, Yang Fan, Kai Dang, Mengfei Du, Xuancheng Ren, Rui Men, Dayiheng Liu, Chang Zhou, Jingren Zhou, and Junyang Lin.
\newblock Qwen2-vl: Enhancing vision-language model's perception of the world at any resolution.
\newblock \emph{arXiv preprint arXiv:2409.12191}, 2024.

\bibitem[Wu et~al.(2023)Wu, Zeng, Wang, and Chen]{wu2023ppt}
Xinjian Wu, Fanhu Zeng, Xiudong Wang, and Xinghao Chen.
\newblock Ppt: Token pruning and pooling for efficient vision transformers.
\newblock \emph{arXiv preprint arXiv:2310.01812}, 2023.

\bibitem[Xing et~al.(2024)Xing, Huang, Dong, Lu, Zhang, Zang, Cao, He, Wang, Wu, et~al.]{xing2024pyramiddrop}
Long Xing, Qidong Huang, Xiaoyi Dong, Jiajie Lu, Pan Zhang, Yuhang Zang, Yuhang Cao, Conghui He, Jiaqi Wang, Feng Wu, et~al.
\newblock Pyramiddrop: Accelerating your large vision-language models via pyramid visual redundancy reduction.
\newblock \emph{arXiv preprint arXiv:2410.17247}, 2024.

\bibitem[Yang et~al.(2024)Yang, Yang, Hui, Zheng, Yu, Zhou, Li, Li, Liu, Huang, Dong, Wei, Lin, Tang, Wang, Yang, Tu, Zhang, Ma, Yang, Xu, Zhou, Bai, He, Lin, Dang, Lu, Chen, Yang, Li, Xue, Ni, Zhang, Wang, Peng, Men, Gao, Lin, Wang, Bai, Tan, Zhu, Li, Liu, Ge, Deng, Zhou, Ren, Zhang, Wei, Ren, Liu, Fan, Yao, Zhang, Wan, Chu, Liu, Cui, Zhang, Guo, and Fan]{yang2024qwen2technicalreport}
An Yang, Baosong Yang, Binyuan Hui, Bo Zheng, Bowen Yu, Chang Zhou, Chengpeng Li, Chengyuan Li, Dayiheng Liu, Fei Huang, Guanting Dong, Haoran Wei, Huan Lin, Jialong Tang, Jialin Wang, Jian Yang, Jianhong Tu, Jianwei Zhang, Jianxin Ma, Jianxin Yang, Jin Xu, Jingren Zhou, Jinze Bai, Jinzheng He, Junyang Lin, Kai Dang, Keming Lu, Keqin Chen, Kexin Yang, Mei Li, Mingfeng Xue, Na Ni, Pei Zhang, Peng Wang, Ru Peng, Rui Men, Ruize Gao, Runji Lin, Shijie Wang, Shuai Bai, Sinan Tan, Tianhang Zhu, Tianhao Li, Tianyu Liu, Wenbin Ge, Xiaodong Deng, Xiaohuan Zhou, Xingzhang Ren, Xinyu Zhang, Xipin Wei, Xuancheng Ren, Xuejing Liu, Yang Fan, Yang Yao, Yichang Zhang, Yu Wan, Yunfei Chu, Yuqiong Liu, Zeyu Cui, Zhenru Zhang, Zhifang Guo, and Zhihao Fan.
\newblock Qwen2 technical report, 2024.

\bibitem[Ye et~al.(2024{\natexlab{a}})Ye, Gan, Ge, Zhang, and Tang]{ye2024atp}
Xubing Ye, Yukang Gan, Yixiao Ge, Xiao-Ping Zhang, and Yansong Tang.
\newblock Atp-llava: Adaptive token pruning for large vision language models.
\newblock \emph{arXiv preprint arXiv:2412.00447}, 2024{\natexlab{a}}.

\bibitem[Ye et~al.(2024{\natexlab{b}})Ye, Gan, Huang, Ge, Shan, and Tang]{ye2024voco}
Xubing Ye, Yukang Gan, Xiaoke Huang, Yixiao Ge, Ying Shan, and Yansong Tang.
\newblock Voco-llama: Towards vision compression with large language models.
\newblock \emph{arXiv preprint arXiv:2406.12275}, 2024{\natexlab{b}}.

\bibitem[Yin et~al.(2022)Yin, Vahdat, Alvarez, Mallya, Kautz, and Molchanov]{Avit}
Hongxu Yin, Arash Vahdat, Jose~M Alvarez, Arun Mallya, Jan Kautz, and Pavlo Molchanov.
\newblock A-vit: Adaptive tokens for efficient vision transformer.
\newblock In \emph{Proceedings of the IEEE/CVF conference on computer vision and pattern recognition}, pages 10809--10818, 2022.

\bibitem[Yu et~al.(2023)Yu, Yang, Li, Wang, Lin, Liu, Wang, and Wang]{yu2023mmvet}
Weihao Yu, Zhengyuan Yang, Linjie Li, Jianfeng Wang, Kevin Lin, Zicheng Liu, Xinchao Wang, and Lijuan Wang.
\newblock {MM-Vet: Evaluating Large Multimodal Models for Integrated Capabilities}.
\newblock \emph{arXiv prepring arXiv:2308.02490}, 2023.

\bibitem[Zhai et~al.(2023)Zhai, Mustafa, Kolesnikov, and Beyer]{siglip}
Xiaohua Zhai, Basil Mustafa, Alexander Kolesnikov, and Lucas Beyer.
\newblock Sigmoid loss for language image pre-training.
\newblock In \emph{Proceedings of the IEEE/CVF international conference on computer vision}, pages 11975--11986, 2023.

\bibitem[Zhang et~al.(2025)Zhang, Fang, Yang, and Feng]{llava_mini}
Shaolei Zhang, Qingkai Fang, Zhe Yang, and Yang Feng.
\newblock Llava-mini: Efficient image and video large multimodal models with one vision token.
\newblock \emph{arXiv preprint arXiv:2501.03895}, 2025.

\bibitem[Zhang et~al.(2024)Zhang, Fan, Ma, Zheng, Huang, Cheng, Gudovskiy, Okuno, Nakata, Keutzer, et~al.]{zhang2024sparsevlm}
Yuan Zhang, Chun-Kai Fan, Junpeng Ma, Wenzhao Zheng, Tao Huang, Kuan Cheng, Denis Gudovskiy, Tomoyuki Okuno, Yohei Nakata, Kurt Keutzer, et~al.
\newblock Sparsevlm: Visual token sparsification for efficient vision-language model inference.
\newblock \emph{arXiv preprint arXiv:2410.04417}, 2024.

\end{thebibliography}
